\documentclass[accepted]{uai2024} 
                        
 \usepackage[british]{babel}

\usepackage{natbib} 
    \renewcommand{\bibsection}{\subsubsection*{References}}
\usepackage{mathtools} 

\usepackage{graphicx}
\usepackage{booktabs}
\usepackage{longtable}
\usepackage{array}
\usepackage{multirow}
\usepackage{wrapfig}
\usepackage{float}
\usepackage{colortbl}
\usepackage{pdflscape}
\usepackage{tabu}
\usepackage{threeparttable}
\usepackage{threeparttablex}
\usepackage[normalem]{ulem}
\usepackage{makecell}
\usepackage{xcolor} 
\usepackage{tikz}
\usepackage{amssymb}

\newcommand{\battery}[2]{%
    \begin{tikzpicture}[scale=1]
        \draw[line width=0.1pt,fill=darkgray!90] (0,0) rectangle (#1,0.35); 
        \draw[line width=0.1pt,fill=gray!20] (#1,0) rectangle (2,0.35); 
        \node[align=center,font=\scriptsize,text=white] at (0.5*#1,0.175) {RF}; 
        \node[align=center,font=\scriptsize] at (1+0.5*#1,0.175) {NRF}; 
    \end{tikzpicture}%
}

\title{Going Further: Flatness at the Rescue of Early Stopping for Adversarial Example Transferability}

\author[1]{\href{https://gubri.eu}{Martin~Gubri}{}\hspace{.15em}\thanks{Work conducted during the author's PhD at the University of Luxembourg.}\hspace{.15em}}
\author[2]{Maxime Cordy}
\author[2]{Yves Le Traon}
\affil[1]{%
    Parameter Lab\\
    Tübingen, Germany
}
\affil[2]{%
    University of Luxembourg\\
    Luxembourg, Luxembourg
}
  
\begin{document}
\maketitle

\begin{abstract}
    Transferability is the property of adversarial examples to be misclassified by other models than the surrogate model for which they were crafted. Previous research has shown that early stopping the training of the surrogate model substantially increases transferability. A common hypothesis to explain this is that deep neural networks (DNNs) first learn robust features, which are more generic, thus a better surrogate. Then, at later epochs, DNNs learn non-robust features, which are more brittle, hence worst surrogate. 
    First, we tend to refute this hypothesis, using transferability as a proxy for representation similarity. We then establish links between transferability and the exploration of the loss landscape in parameter space, focusing on sharpness, which is affected by early stopping. This leads us to evaluate surrogate models trained with seven minimizers that minimize both loss value and loss sharpness. Among them, SAM consistently outperforms early stopping by up to 28.8 percentage points. We discover that the strong SAM regularization from large flat neighborhoods tightly links to transferability. Finally, the best sharpness-aware minimizers prove competitive with other training methods and complement existing transferability techniques.
\end{abstract}

\section{Introduction}
\label{sec:intro}

\begin{figure}[t]
\begin{center}
   \includegraphics[width=0.85\linewidth]{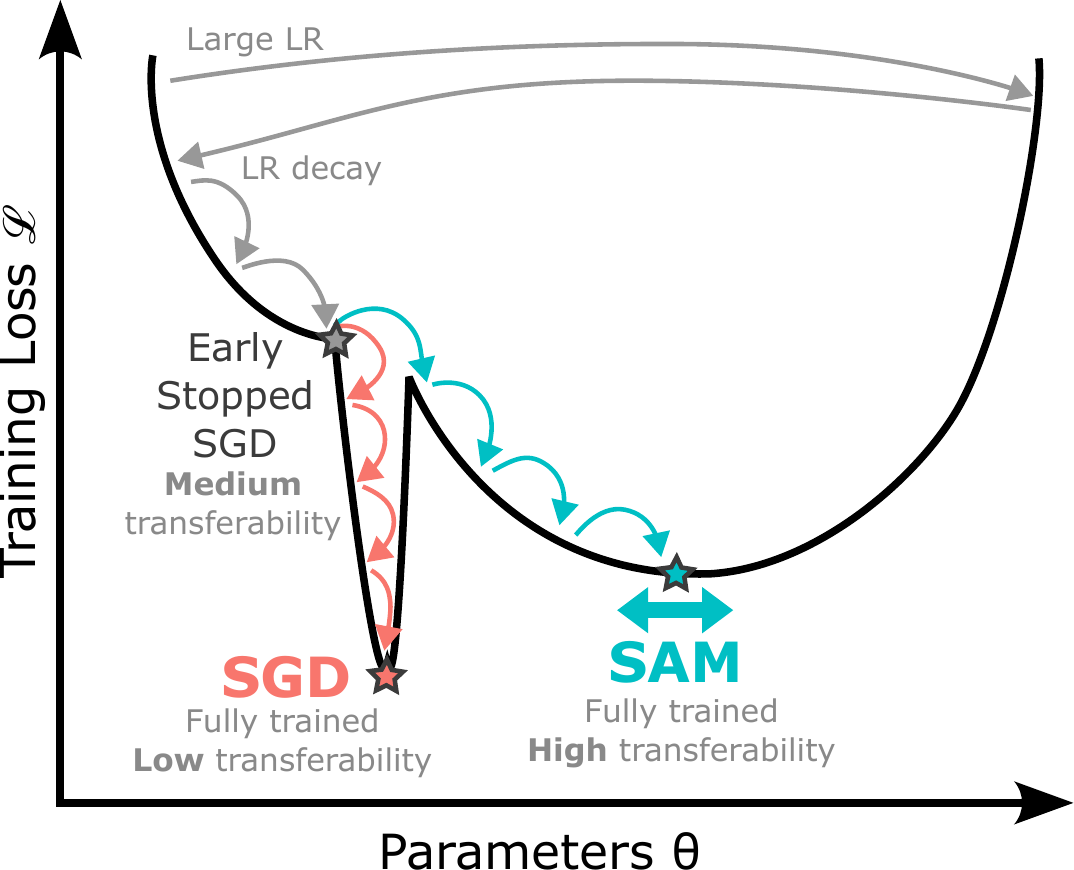}
\end{center}
   \caption{Illustration of the loss landscape, showing the training of surrogate models to craft transferable adversarial examples. Before the learning rate decays, training tends to ``cross the valley'' with plateauing transferability. A few iterations after the decay of the learning rate, early stopped SGD achieves its best transferability (gray). In the following epochs, SGD falls progressively into deep, sharp holes in the parameter space with poor transferability (red). l-SAM (blue) avoids these holes by minimizing the maximum loss around an unusually large neighborhood (thick blue arrow).}
\label{fig:illustration}
\end{figure}

State-of-the-art Deep Neural Networks (DNNs) are vulnerable to imperceptible worst-case inputs perturbations, so-called adversarial examples \citep{Biggio2013,Szegedy2013}. These perturbations are not simple flukes of specific representations because some are simultaneously adversarial against several independently trained models with distinct architectures \citep{Goodfellow2014ExplainingExamples}. This observation leads to the discovery of the \textit{transferability} of adversarial examples, i.e., an adversarial example against a model is likely to be adversarial against another model. This phenomenon is not well understood but has practical implications. Indeed, practitioners cannot rely on security by obscurity. Attackers can apply white-box attacks to their \textit{surrogate model} to fool an unknown \textit{target model}. These types of attack are called transfer-based back-box attacks. They do not require \textit{any} query access to the model to craft adversarial examples. Crafting highly transferable adversarial examples for distinct architectures is still an open problem \citep{Naseer2022OnTransformers} and an active area of research \citep{Benz2021BatchPerspective,Dong2018BoostingMomentum,Gubri2022EfficientNetworks,Gubri2022LGV:Vicinity,Li2018LearningNetworks,Lin2019,Springer2021AAttacks,Wu2020SkipResNets,Xie2018,Zhao2022TowardsAttacks}. Understanding the underlining characteristics that drive transferability provides insights into how DNNs learn generic representations. 

Despite strong interest in transferability, little attention has been paid to how to train better surrogate models. The most commonly used method is arguably \emph{early stopping} \citep{Benz2021BatchPerspective,Zhang2021EarlyTransferability,Nitin2021SGDNon-Robust} -- which is originally a practice to improve natural generalization. The commonly accepted hypothesis to explain why early stopping improves transferability is that an early stopped DNN is composed of more robust features, whereas the fully trained counterpart has more brittle non-robust features \citep{Benz2021BatchPerspective,Zhang2021EarlyTransferability,Nitin2021SGDNon-Robust}. 

In this paper, we invalidate this hypothesis empirically and uncover other explanations behind the effectiveness of early stopping, and more generally on how to achieve better surrogate training. We observe in Section~\ref{sec:early-stopping} that early stopping also improves transferability from and to models composed of \emph{non-robust} features. We propose an alternative hypothesis centered on the dynamics of loss surface exploration, with a focus on sharpness. Section~\ref{sec:training-dynamics} motivates this approach, highlighting that transferability peaks when the loss sharpness in the weight space drops. Section~\ref{sec:sam} shows that seven optimizers significantly increase the transferability of a surrogate model by minimizing its sharpness. In particular, we reveal that the stronger regularization induced by Sharpness-Aware Minimizer (SAM) with unusually large neighborhood (l-SAM), improves transferability specifically, since l-SAM and SGD have a similar natural generalization. We conclude that this strong regularization alters the exploration of the loss landscape by avoiding deep, sharp holes where the learned representation is too specific. Finally, Section~\ref{sec:evalution} evaluates l-SAM and two variants against other surrogate training procedures, and combined with nine non-training transferability techniques. 

Figure \ref{fig:illustration} illustrates the insights and grounded principles to improve transferability that our contribution brings:
\begin{itemize}
    \item The learning rate decay allows the exploration of the loss landscape to go down the valley. After a few iterations, SGD reaches its best transferability (``early stopped SGD'', gray star). The sharpness is moderate.
    \item As training with SGD continues, sharpness increases and transferability decreases. The fully trained model (red star) is a suboptimal surrogate. SGD falls into deep, sharp holes where the representation is too specific. 
    \item SAM explicitly minimizes sharpness and avoids undesirable holes. Transferability is maximum after a full training (blue star) when SAM is applied over a large neighborhood (l-SAM, thick blue arrow). 
\end{itemize}

\section{Related Work}
\label{sec:related}

\paragraph{Transferability techniques.}
The transferability of adversarial examples is a prolific research topic \citep{Benz2021BatchPerspective,Dong2018BoostingMomentum,Gubri2022EfficientNetworks,Gubri2022LGV:Vicinity,Li2018LearningNetworks,Lin2019,Springer2021AAttacks,Wu2020SkipResNets,Xie2018,Zhao2022TowardsAttacks}. \citet{Zhao2022TowardsAttacks} suggest comparing transferability techniques within specific categories, a recommendation adhered to in Section \ref{sec:evalution}. They classify gradient-based transferability techniques into four categories: model augmentation, data augmentation, attack optimizers, and feature-based attacks. Section \ref{sec:evalution} shows that our method improves these techniques when combined. Model augmentation adds randomness to the weights or the architecture to avoid specific adversarial examples: GN \citep{Li2018LearningNetworks} uses dropout or skip erosion, SGM \citep{Wu2020SkipResNets} favors gradients from skip connections during the backward pass, LGV \citep{Gubri2022LGV:Vicinity} collects models along the SGD trajectory during a few additional epochs with a high learning rate. Data augmentation techniques transform the inputs during the attack: DI \citep{Xie2018} randomly resizes the input, SI \citep{Lin2019} rescales the input, and VT \citep{Wang2021EnhancingTuning} smooths the gradients locally. Attack optimizers smooth updates during gradient ascent with momentum (MI, \citet{Dong2018BoostingMomentum}) or Nesterov accelerated gradient (NI, \citet{Lin2019}), or minimize sharpness (RAP, \citep{qinBoostingTransferabilityAdversarial2022}). RAP minimizes sharpness through a min-max bi-level optimization problem, similar to SAM but in the input space. Section \ref{sec:evalution} shows that SAM and RAP are best combined, indicating their complementary effects on two distinct factors.

\paragraph{Training surrogate models.} 
Despite the important amount of work on transferability, the way to train an effective single surrogate base model has received little attention in the literature \citep{Zhao2022TowardsAttacks}. \citet{Benz2021BatchPerspective,Nitin2021SGDNon-Robust,Zhang2021EarlyTransferability} point that early stopping SGD improves transferability. \citet{Springer2021AAttacks} propose SAT, slight adversarial training that uses tiny perturbations to filter out some non-robust features. 
Section~\ref{sec:sam} evaluates SAT. Our approach sheds new light on the relation between flatness and transferability. \citet{Springer2021AAttacks} implicitly flatten the surrogate model, since adversarial trained models are flatter than their naturally trained counterparts \citep{Stutz2021RelatingMinima}. We observe a similar implicit link with early stopping in Section~\ref{sec:training-dynamics}. \citet{Gubri2022LGV:Vicinity} propose the surrogate-target misalignment hypothesis to explain why flat minima in the weight space are better surrogate models. Section \ref{sec:evalution} shows that LGV, their model augmentation technique, is best with ours, indicating complementary effects on two distinct factors, respectively ensembling diverse representations and training a single generic representation. 

\paragraph{Early stopping for transferability.} \citet{Benz2021BatchPerspective,Zhang2021EarlyTransferability,Nitin2021SGDNon-Robust} point out that fully trained surrogate models are not optimal for transferability and propose a hypothesis based on the perspective of robust and non-robust features \citep{Ilyas2019AdversarialFeatures}. \citet{Ilyas2019AdversarialFeatures} disentangles features that are highly predictive and robust to adversarial perturbations (RFs), and features that are also highly predictive but non-robust to adversarial perturbations (NRFs). A feature $f_r$ is $\gamma$-robust if it remains predictive under a specified set of adversarial perturbations $\Delta$, $\mathbb{E}_{(x,y)\sim D} \left[ \inf_{\delta \in \Delta(x)} y \cdot f_r(x + \delta) \right] \geq \gamma$. A NRF $f_{nr}$ is $\beta$-predictive, i.e, $\mathbb{E}_{(x,y)\sim D} \left[ y \cdot f_{nr}(x) \right] \geq \beta$ but is not $\gamma$-robust feature for any $\gamma \geq 0$. According to \citet{Benz2021BatchPerspective,Nitin2021SGDNon-Robust}, the training of DNNs mainly learns RFs first and then learns NRFs. We term this the \textit{RFs/NRFs evolution hypothesis} (Table \ref{tab:rfs-nrfs-hypothesis}). NRFs are transferable \citep{Ilyas2019AdversarialFeatures}, but also brittle. RFs are more stable and can improve transferability \citep{Springer2021AAttacks,Zhang2021EarlyTransferability}. RFs can be attacked: RFs against $L_p$ norm $\varepsilon$ perturbations are vulnerable against perturbations with higher $\varepsilon'$ or another $L_{p'}$ norm \citep{Springer2021AAttacks,Zhang2021EarlyTransferability}. Section~\ref{sec:early-stopping} provides some observations that tend to refute the RFs/NRFs evolution hypothesis (summary in Table \ref{tab:rfs-nrfs-summary}). 

\paragraph{Sharpness and natural generalization.}

Several training techniques increase natural generalization and reduce loss sharpness in the weight space. \citet{keskar2017on} links batch size to sharpness, defined in the weight space as $\max_{\|\epsilon\|_2 \leq \rho} \mathcal{L}(w + \epsilon) - \mathcal{L}(w)$. SWA \citep{izmailovAveragingWeightsLeads2018} averages the weights at the last epochs for flatness. SAM \citep{Foret2020Sharpness-AwareGeneralization} minimizes the maximum loss around a neighborhood by performing a gradient ascent step, defined by $\epsilon_t = \rho \frac{\nabla \mathcal{L}(w_t)}{\|\nabla \mathcal{L}(w_t)\|_2}$, followed by a gradient descent step, $w_{t+1} = w_t - \alpha_t \left( \nabla \mathcal{L}(w_t + \epsilon_t) + \lambda w_t \right)$. At the cost of one additional forward-backward pass per iteration, SAM avoids deep, sharp holes on the surface of the loss landscape \citep{Kaddour2022WhenWork}. Its $\rho$ hyperparameter controls the size of flat neighborhoods. Several variants exist that improve natural generalization \citep{Kwon2021ASAM:Networks,Zhuang2022SurrogateTraining} or efficiency \citep{liuEfficientScalableSharpnessAware2022,duEfficientSharpnessawareMinimization2021} (description in Appendix~\ref{sec:app-sam-hp}). Nevertheless, the relationship between sharpness and natural generalization is subject to scientific controversy \citep{andriushchenkoModernLookRelationship2023,wenSharpnessMinimizationAlgorithms2023,pmlr-v151-bisla22a}. In Sections \ref{sec:sam} and \ref{sec:evalution}, we explore the use of SWA, SAM and six variants to train better surrogate models.

\section{Another Look at the Non-Robust Features Hypothesis} 
\label{sec:early-stopping}

\begin{table}[t]
\centering
\caption{RFs/NRFs evolution hypothesis (expected) and hypothesis based on our results (ours), about the evolution of DNN features from early stopping to full training.}
\label{tab:rfs-nrfs-hypothesis}
\begin{tabular}{@{}lll@{}}
\toprule
         & \multicolumn{2}{c}{Evolution of DNN features} \\ \cmidrule(l){2-3}
Prev. hypothesis &   \battery{1.53}{} \hspace{-1.35em}  & $\xrightarrow[]{\text{training}}$ \battery{0.44}{}     \\
Ours     &   \battery{1.00}{} \hspace{-1.35em}  & $\xrightarrow[]{\text{training}}$ \battery{1.00}{}     \\
\bottomrule
\end{tabular}
\end{table}

\begin{table}[t]
\centering
\caption{Comparison of expected and observed evolutions of transferability from early stopped to fully trained \textit{surrogate} model. DNN denotes a regularly trained model.}
\label{tab:rfs-nrfs-summary}
\begin{tabular}{@{}llll@{}}
\toprule
      &  Transferability  & \multicolumn{2}{c}{\small Evolution of transferability} \\ \cmidrule(l){3-4}
      &  surrogate $\mapsto$ target              & Expected                                                 & Observed                                                 \\ \midrule
    \multirow{2}{*}{Fig. \ref{fig:xp_rfs_nrfs_plot_dr_dnr_agg}} & RFs \enspace $\, \mapsto$ DNN                                                                     & $\searrow$                                                 & $\searrow$ {\scriptsize(blue)}                                                   \\
      & NRFs $\mapsto$ DNN                                     & $\nearrow$                                             & $\searrow$ {\scriptsize(green)}                                                   \\ \midrule 
    \multirow{2}{*}{Fig. \ref{fig:xp_rfs_nrfs_plot_from_d_to_d_modified}} & DNN $\mapsto$ RFs                                                                      & $\searrow$                                                & $\searrow$ {\scriptsize(red)}                                                   \\
      & DNN $\mapsto$ NRFs                                                                     & $\nearrow$                                                 & $\searrow$ {\scriptsize(others)}                                                   \\ 
\bottomrule
\end{tabular}
\end{table}

In this section, we point the flaws of the robust and non-robust features~(RFs/NRFs) evolution hypothesis \cite{Benz2021BatchPerspective,Zhang2021EarlyTransferability,Nitin2021SGDNon-Robust} to explain the success of early stopping for transferability. They observe that early stopping the surrogate model improves the transferability. To explain this, they draw the hypothesis that DNNs first mainly learn RFs and then NRFs (Table \ref{tab:rfs-nrfs-hypothesis}). Since RFs are less brittle than NRFs, early learned features would be more transferable than their fully trained counterparts \footnote{Adversarial perturbations against RFs exist: either perturbations with larger $L_p$ norms $\varepsilon$, or with another $L_{p'}$ norm, as chosen by \citet{Springer2021AAttacks,Zhang2021EarlyTransferability}.}. We challenge the fact that DNNs first mainly learn RFs and then NRFs. Considering transferability as a proxy for representation similarity, we show no trade-off between RFs and NRFs along training epochs (see summary in Table \ref{tab:rfs-nrfs-summary}). First, the transferability from RFs to a regular DNN evolve similarly as the transferability from NFRs to a regular DNN. Second, the transferabilities from a regular surrogate DNN to both RFs and NRFs target DNNs evolve similarly. 

\paragraph{Early stopping indeed increases transferability.} First, we check that a fully trained surrogate model is not optimal for transferability. We train two ResNet-50 surrogate models on CIFAR-10 and ImageNet using standard settings. Appendix \ref{sec:app-transferability-epochs} reports the success rates of the BIM attack applied at every epoch and evaluated on 10 fully trained target models per dataset. For both datasets and diverse targeted architectures, the optimal epoch for transferability occurs around one or two thirds of training\footnote{Transferability decreases along epochs, except for the two vision transformers targets on ImageNet where the transferability is stable at the end of training.}.

\paragraph{Early stopping improves transferability from both surrogates trained on robust and non-robust features.} We show that early stopping works \textit{similarly} well on surrogate models trained on robust and non-robust datasets. We retrieve the robust and non-robust datasets from \cite{Ilyas2019AdversarialFeatures}, that are altered from CIFAR-10 to mostly contain RFs and, respectively, NRFs. We train two ResNet-50 models on both datasets with SGD (hyperparameters reported in Appendices B and C). Figure~\ref{fig:xp_rfs_nrfs_plot_dr_dnr_agg} shows the transferability across training epochs, averaged over the ten regularly trained targets. The success rates of both robust and non-robust surrogate models evolve similarly to the model trained on the original dataset: transferability peaks around the epochs 50 and 100 and decreases during the following epochs. This observation is valid for all ten targets (Appendix \ref{sec:app-rfs-nrfs}). According to the RFs/NRFs evolution hypothesis, we expect ``X-shaped'' transferability curves: if DNNs first mainly learn RFs and then NRFs, the transferability from NRFs would increase and the transferability from RFs would strictly decrease (from early stopping to full training). The RFs/NRFs hypothesis does not describe why early learned NRFs are better to target a regular DNN than fully learned NRFs.

\begin{figure}[t]
    \centering
    \includegraphics[width=0.9\linewidth]{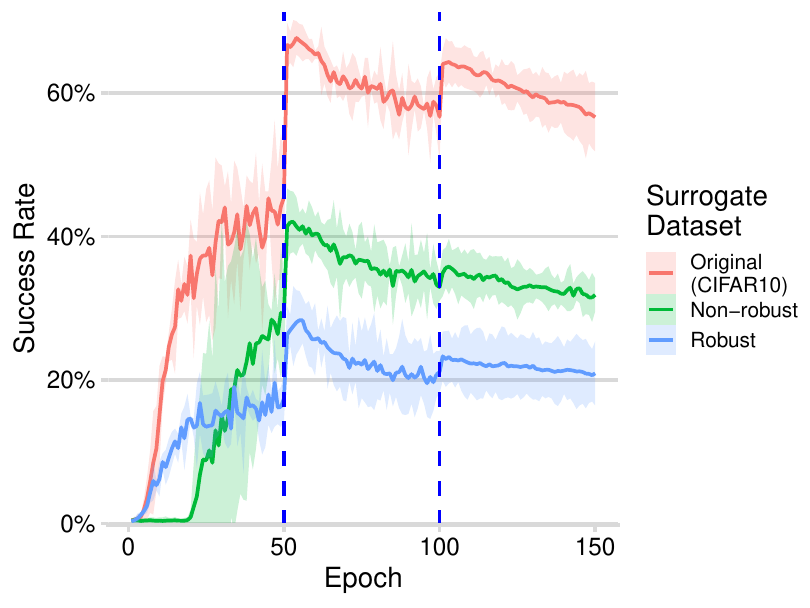}
    \caption{Early stopping improves the transferability \textit{from} surrogate models trained on both robust and non-robust datasets. Average success rate evaluated over ten target models trained on the original CIFAR-10 dataset, from a ResNet-50 surrogate model trained for a number of epochs (x-axis) on the datasets $D_R$ (blue) and $D_\text{NR}$ (green) of \cite{Ilyas2019AdversarialFeatures} modified from CIFAR-10 (red). We craft all adversarial examples from the same subset of the original CIFAR-10 test set. Average (line) and confidence interval of $\pm$ two standard deviations (colored area) of three training runs. Appendix \ref{sec:app-rfs-nrfs} contains the details per target.}
    \label{fig:xp_rfs_nrfs_plot_dr_dnr_agg}
\end{figure}

\begin{figure}[t]
    \centering
    \includegraphics[width=0.9\linewidth]{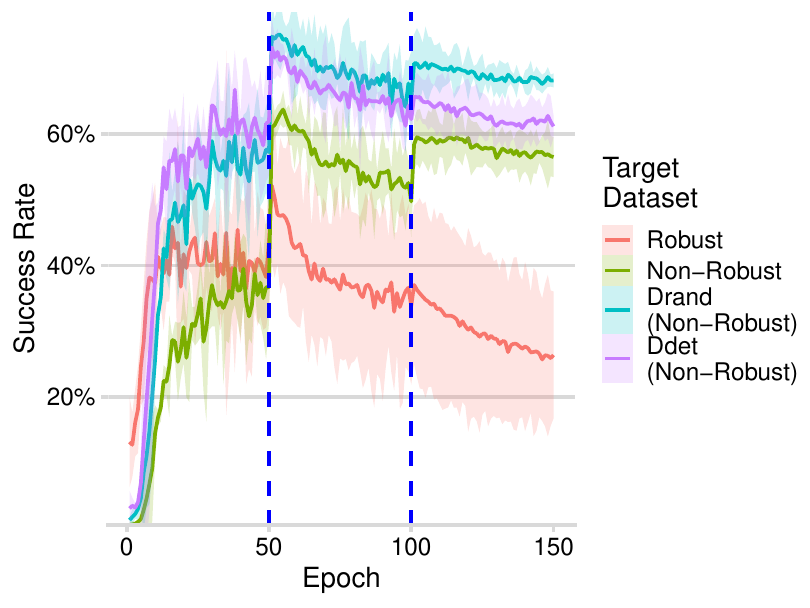}
    \caption{Early stopping improves the transferability \textit{to} target models trained on both robust and non-robust datasets. Success rate from a ResNet-50 trained for a number of epochs (x-axis) on the original CIFAR-10 dataset, to ResNet-50 targets trained on the robust dataset $D_R$ (red), and the three non-robust datasets $D_\text{NR}$ (green), $D_\text{rand}$ (blue) and $D_\text{det}$ (purple) of \citet{Ilyas2019AdversarialFeatures} modified from CIFAR-10. The perturbation norm $\varepsilon$ is $\frac{16}{255}$ for the $D_R$ target, $\frac{2}{255}$ for the $D_{NR}$ target and $\frac{1}{255}$ for the $D_\text{rand}$ and $D_\text{det}$ targets to adapt to the vulnerability of target models (the order of lines cannot be compared). Average (line) and confidence interval of $\pm$ two standard deviations (colored area) of three training runs.}
    \label{fig:xp_rfs_nrfs_plot_from_d_to_d_modified}
\end{figure}

\paragraph{Early stopping improves transferability to both targets trained on robust and non-robust features.} We observe that an early stopped surrogate model trained on the original dataset is best to target both RFs and NRFs targets. Here, we keep the original CIFAR-10 dataset to train the surrogate model. We target four ResNet-50 models trained on the robust and non-robust datasets of \cite{Ilyas2019AdversarialFeatures}\footnote{In this experiment, we include two additional non-robust datasets $D_\text{rand}$ and $D_\text{det}$ from \cite{Ilyas2019AdversarialFeatures}. By construction, their \textit{only} useful features for classification are NRFs. They were excluded from the previous experiment due to training instability.}. Figure~\ref{fig:xp_rfs_nrfs_plot_from_d_to_d_modified} shows that the same epoch of standard training is optimal for attacking all four models, i.e., composed of either RFs or NRFs. The RFs/NRFs evolution hypothesis fails to explain why early stopping is best to target NRFs.

Overall, we provide new evidence that early stopping for transferability acts similarly on robust and non-robust features. We do not observe an inherent trade-off between RFs and NRFs. Since the higher the transferability, the more similar the representations are, we conclude that the early trained representations are more similar to both RFs and NRFs than their fully trained counterparts. Therefore, the hypothesis that early stopping favors RFs over NRFs does not hold. We conjecture that a phenomenon orthogonal to RFs/NRFs explains why fully trained surrogates are not optimal.

\section{Stopping Earlier: Transferability and Training Dynamics}
\label{sec:training-dynamics}
This section investigates the link between surrogate model training dynamics and transferability, highlighting that transferability peaks when sharpness drops.

\begin{figure}[t]
    \centering
    \includegraphics[width=0.9\linewidth]{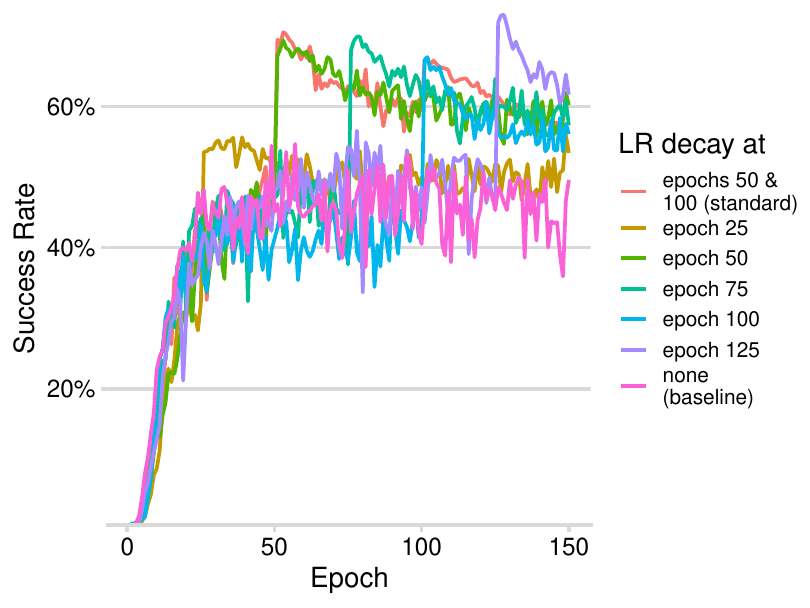}
    \caption{Transferability peaks when the learning rate decays at any epochs. Average success rate evaluated over ten target models from a ResNet-50 surrogate model trained for a number of epochs (x-axis) on CIFAR-10. The learning rate is divided by 10 once during training at the epoch corresponding to the color. Red is our standard schedule, with two decays at epochs 50 and 100. Pink is the baseline of constant learning rate. Best seen in colors.}
        \label{fig:xp_lr_single_decay_plot_successrate_epochs_single_decay_agg}
\end{figure}

\paragraph{Transferability peaks when the LR decays.} 
The optimal number of surrogate training epochs for transferability occurs just after the decay of the LR. We train a ResNet-50 surrogate model for 150 epochs on CIFAR-10, using the standard LR schedule of \cite{Engstrom2019RobustnessLibrary} which divides the LR by 10 at epochs 50 and 100. For the ten targets considered individually, the highest transferability is between epochs 51 and 55 (Appendix B). Figure \ref{fig:xp_lr_single_decay_plot_successrate_epochs_single_decay_agg} shows that transferability suddenly peaks after both LR decays (red line). We observe the same phenomenon on ImageNet\footnote{On ImageNet, we train a ResNet-50 surrogate for 90 epochs with LR decays at epochs 30 and 60. The highest transferability per target occurs either after the first decay (epochs 31 or 35) or after the second one (epochs 62 or 67), except for both vision transformer targets, where transferability plateaus at a low success rate after the second decay.}. Overall, the success of early stopping appears to be related to the exploration of the loss landscape, which is governed by the learning rate.

\paragraph{Consistency of the peak of transferability across training.}

This peak of transferability can be consistently observed at any point of training. Here, we modify the standard double decay LR schedule to perform a single decay at a specified epoch. The learning rate is constant (0.1) until the specified epoch, where it is ten times lower for the rest of the training. We evaluate the transferability of five surrogates with a decay at, respectively, epoch 25, 50, 75, 100 and 125. Figure \ref{fig:xp_lr_single_decay_plot_successrate_epochs_single_decay_agg} reports a similar transferability peak for all these surrogates, except for the smaller peak at epoch 25 where the decay occurs before the end of the initial convergence. The consistency of the peak of transferability across training epochs is valid for all individual targets (Appendix~D). As a baseline, we use a constant learning rate, under which transferability plateaus without any observed peak. Therefore, we conclude that the step decay of the LR enables early stopping to improve transferability.

\begin{figure}[t]
    \centering
    \includegraphics[width=0.9\linewidth]{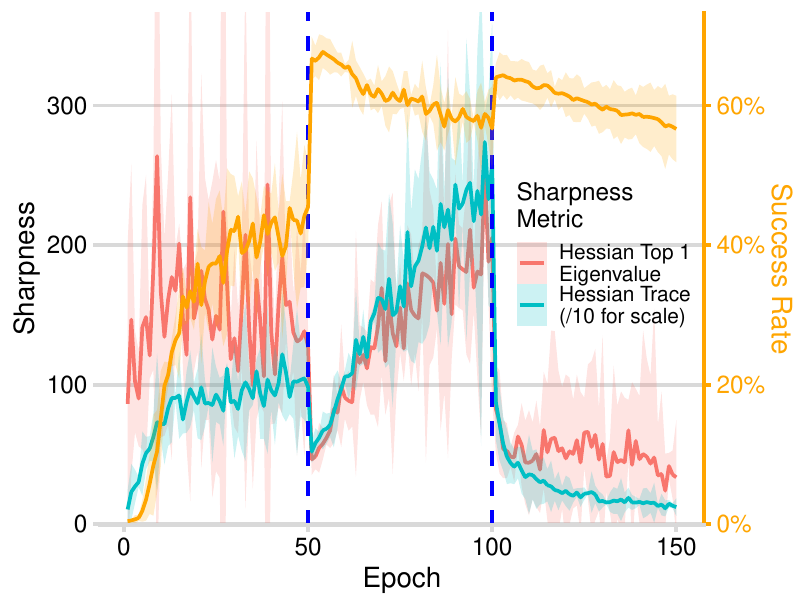}
    \caption{Sharpness drops when the learning rate decays. Largest eigenvalue of the Hessian (red) and trace of the Hessian (blue) for all training epochs (x-axis) on CIFAR-10. Average success rate on ten targets (orange, right axis). Average (line) and confidence interval of $\pm$ two standard deviations (colored area) of three training runs. Vertical bars indicate the learning rate step decays. Best seen in colors.}
    \label{fig:plot_hessian_epochs_final}
\end{figure}

\paragraph{Sharpness drops when the LR decays.}

When the LR decays, the sharpness in the weight space drops. Figure~\ref{fig:plot_hessian_epochs_final} reports two sharpness metrics per the training epoch of our standard CIFAR-10 surrogate: the largest Hessian eigenvalue  measures the sharpness of the sharpest direction in the weight space (red, worst-case sharpness) and the Hessian trace measures the total sharpness of all weight space directions (blue, average sharpness). Both types of sharpness decrease abruptly, significantly and immediately after both LR decays. Simultaneously, transferability peaks (orange).

We conclude that the effect of early stopping on transferability is tightly related to the dynamics of the exploration of the loss surface, governed by the learning rate. Overall, Figure \ref{fig:illustration} illustrates our observations:
\begin{enumerate}
    \item Before the LR decays, the training bounces back and forth crossing the valley from above (top gray arrows). For an extended discussion, see Appendix \ref{sec:app-training-dynamics}.
    \item After the LR decays, training goes down the valley. Soon after, SGD has its best transferability (``early stopped SGD'' gray star). Sharpness is reduced.
    \item As training continues, the loss decreases while sharpness slowly increases. SGD settles into a ``deep hole'' in the loss landscape, with specific representations of low transferability (``fully trained SGD'' red star).
\end{enumerate}

\section{Going Further: Flatness at the Rescue of SGD}
\label{sec:sam}

Since transferability peaks to its higher value when sharpness drops, we now explore how to improve transferability by minimizing the sharpness of the surrogate model. First, we show that seven training techniques that minimize both the loss value and the loss sharpness can train better surrogate models. Second, we uncover that SAM (and five variants) with unusually large flat neighborhoods induces a stronger regularization that specifically increases transferability.

\begin{figure*}[t]
    \begin{center}
        \includegraphics[width=0.8\linewidth]{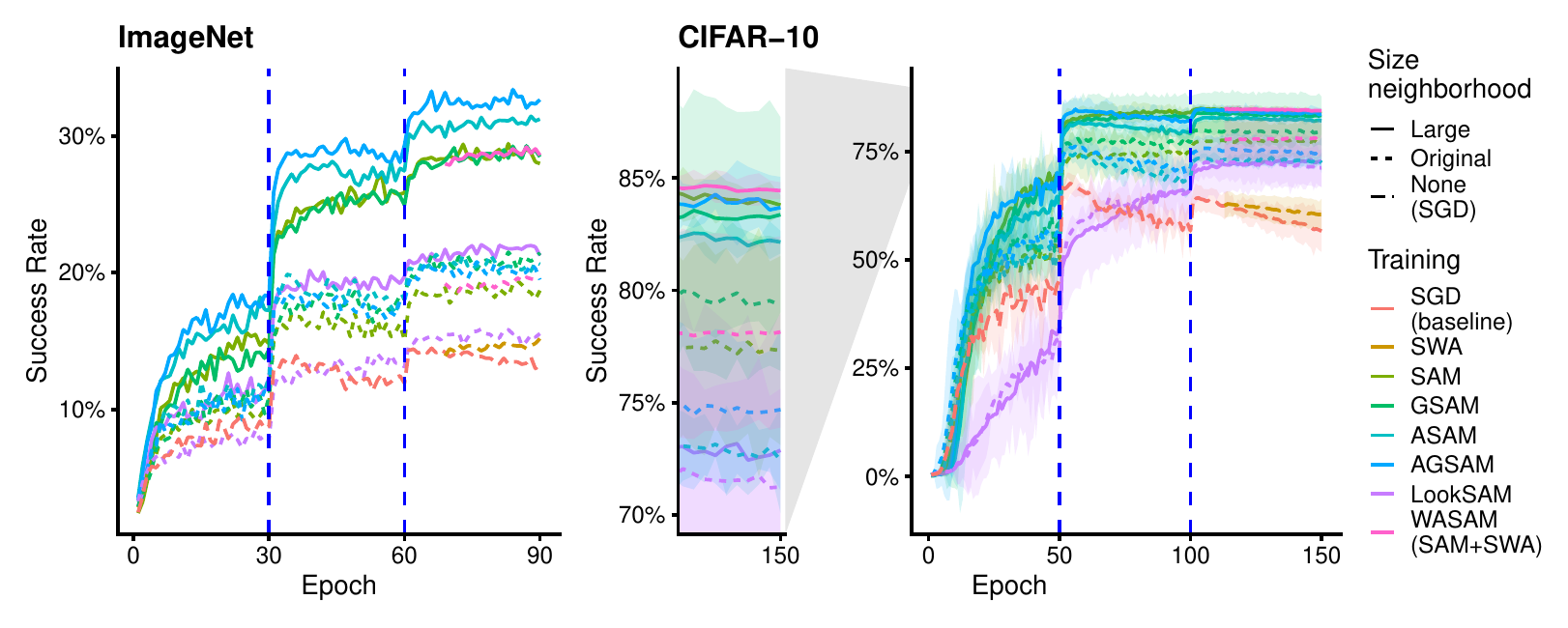}
    \end{center}
   \caption{SAM variants and SWA improve transferability over SGD, and SAM with large neighborhoods over the original SAM. Average success rate evaluated over ten target models from a ResNet-18 surrogate model trained for a number of epochs (x-axis) on ImageNet (left), and from a ResNet-50 on CIFAR-10 (right). SAM and its variants are trained with both the original size of flat neighborhood (dotted, $\rho=0.05$ except $\rho=0.5$ for adaptive variants) and the larger size that we tuned for transferability (plain). Red is our standard SGD surrogate. Best seen in colors.}
    \label{fig:xp_sam_plot_successrate_epochs_agg}
\end{figure*}

    \paragraph{Minimizing sharpness improves transferability.}

The training techniques known to decrease the sharpness of the models train better surrogate representations. We evaluate the transferability of seven training techniques belonging to two families, SWA and SAM (see Section \ref{sec:related} and Appendix~\ref{sec:app-xp-settings} for a more detailed presentation). SWA \citep{izmailovAveragingWeightsLeads2018} decreases sharpness implicitly by averaging the weights collected by SGD. Our SWA surrogate is the average of the weights obtained by our standard SGD surrogate at the end of the last 25\% epochs\footnote{We also update the batch-normalization statistics of the SWA model with one forward pass over the training data on CIFAR-10 (10\% on ImageNet).}. Figure \ref{fig:xp_sam_plot_successrate_epochs_agg} shows that SWA (yellow) improves the success rate compared to fully trained SGD (red) on both datasets. On ImageNet, SWA beats the early stopped SGD surrogate, but not on CIFAR-10. Indeed, SWA helps to find flatter solutions than those found by SGD, but SWA is confined to the same basin of attraction \citep{Kaddour2022WhenWork}. To remediate to this issue, we also train several surrogate models with SAM \citep{Foret2020Sharpness-AwareGeneralization} and its variants, i.e., GSAM \citep{Zhuang2022SurrogateTraining}, ASAM \citep{Kwon2021ASAM:Networks}, AGSAM (GSAM+ASAM), WASAM (SAM+SWA, \citet{Kaddour2022WhenWork}), and LookSAM \citep{liuEfficientScalableSharpnessAware2022}. 
SAM explicitly minimizes sharpness during training by solving a min-max optimization problem. At each iteration step, SAM first maximizes the loss in a neighborhood to compute a second gradient that is used to minimize the loss (details in Appendix \ref{sec:app-sam-hp}). We train one model per SAM variant using the original SAM hyperparameter ($\rho=0.05$). Figure~\ref{fig:xp_sam_plot_successrate_epochs_agg} shows that SAM and its variants (dotted lines) train surrogate models that have a significantly higher transferability than fully trained SGD, early stopped SGD and SWA, on both datasets. On ImageNet, the success rate of SAM averages over the ten targets at 18.7\%, compared to 13.3\% for full training with SGD, 14.5\% for SGD at its best (epoch 66) and 15.2\% for SWA, and respectively 77.3\%, 56.6\%, 67.7\% (epoch 54) and 60.5\% on CIFAR-10. SAM finds different basins of attractions than SGD \citep{Foret2020Sharpness-AwareGeneralization,Kaddour2022WhenWork}. Therefore, some basins of attraction are better surrogate than others, and explicitly minimizing sharpness reaches better ones.

    \paragraph{Strong regularization from large flat neighborhoods significantly improves transferability.}

We uncover that the size of the flat neighborhood of SAM and its variants induces a regularization that is tightly linked to transferability. We observe that SAM and its variants with uncommonly large flat neighborhoods train significantly and consistently better surrogate models. SAM seeks neighborhoods with uniformly low loss of size controlled by its $\rho$ hyperparameter (Section \ref{sec:related}). We tune it on CIFAR-10 on distinct validation sets of natural examples, target, and surrogate models (details in Appendix \ref{sec:app-sam-hp}). For all SAM variants, the optimal $\rho$ for transferability is always larger than the original $\rho$, and unusually large compared to the range of values used for natural accuracy. Indeed, we find $\rho$ of 0.3 optimal for SAM, and \cite{Foret2020Sharpness-AwareGeneralization} originally uses $\rho$ of 0.05. \cite{Kaddour2022WhenWork} and \cite{Zhuang2022SurrogateTraining} tune $\rho$ with a maximum of, respectively, 0.2 and 0.3. Figure~\ref{fig:xp_sam_plot_successrate_epochs_agg} reports the transferability of SAM and its variants with both the original $\rho$ (dotted) and with the larger $\rho$ found optimal on CIFAR-10 (plain). All SAM variants train a better surrogate model with large $\rho$ values\footnote{As expected, LookSAM shows a slower learning behavior over the epoch, compared to other SAM variants: LookSAM is an efficient variant that computes the additional SAM gradient only once each five optimizer iteration.}. In the following, we denote \textbf{l-SAM} for SAM with large $\rho$ (0.3), and similarly \textbf{l-AGSAM} and \textbf{l-LookSAM} (respectively, 4\footnote{l-AGSAM uses $\rho$ value of 4, since as suggested by \cite{Kwon2021ASAM:Networks} adaptive variants should use $\rho$ 10 times larger. We found our observations consistent with this recommandation.} and 0.3). \cite{Kaddour2022WhenWork} show that changing $\rho$ ends up in different basins of attraction. Therefore, the stronger regularization induced by l-SAM avoids large sharp holes on top of the loss surface, and significantly improves transferability.

    \paragraph{The benefits of the strong regularization from large flat neighborhoods are specific to transferability.} 

The stronger regularization of SAM with a large value of $\rho$ is specifically related to transferability. First, this strength of regularization may degrade natural accuracy. On ImageNet with ResNet-18, the top-1 accuracy of SAM with large $\rho$ is equal to 67.89\%, less than SAM with the original $\rho$ (70.29\%) and even less than fully trained SGD (69.84\%). This observation extends to ResNet-50 and to the other variants of SAM on ImageNet (Appendix \ref{sec:app-sam-hp}). Therefore, the improvement in generalization of adversarial examples cannot be explained by an improvement in natural generalization (better fit to the data). Second, unlike SAM, a stronger regularization of weight decay decreases transferability, showing a specific relation between transferability and SAM. We train multiple surrogate models using SGD with different values of weight decay. The optimal weight decay value for the ResNet-50 surrogate is the same value used to train the target model (see Appendix~\ref{sec:app-wd} for details). Therefore, not all regularization schemes help to train a better surrogate model. 

Overall, the sharpness of the surrogate model is tightly related to transferability:
\begin{itemize}
    \item Minimizing implicitly or explicitly the loss sharpness trains better surrogate models.
    \item The strong SAM regularization avoids deep sharp minima in favor of unusually large flat neighborhoods that contain more generic representations.
    \item The stronger SAM regularization is tailored for transferability: it can reduce natural accuracy, and other strong regularization schemes, such as weight decay, do not aid in transferability.
\end{itemize}

\begin{table*}[t]
\begin{center}
\caption{Success rate and computation cost of surrogate training techniques on ImageNet and CIFAR-10. Average success rate on ten targets from a ResNet-50 surrogate with a maximum perturbation $L_\infty$ norm $\varepsilon$ of $4/255$ (other norms in Appendix). The computational overhead is relative to the number of SGD forward-backward passes. Bold is best. In \%.}
\label{tab:competitive-ttechs-new}
\begin{tabular}{@{}lrrllrrrr@{}}
\toprule
 &
  \multicolumn{2}{c}{Success Rate $\uparrow$} &
  \multicolumn{2}{c}{Computation Cost $\downarrow$} \\ 
  \cmidrule(l{3pt}r{3pt}){2-3} 
  \cmidrule(l{3pt}r{3pt}){4-5} 
Surrogate &
  ImageNet &
  CIFAR-10 &
  ImageNet &
  CIFAR-10 \\ \midrule
\rowcolor{gray!6} 
Fully Trained SGD & 17.81          & 56.06          & $\times$ 1    & $\times$ 1    \\
Early Stopped SGD & 19.97          & 70.16          & $\times$ 0.77 & $\times$ 0.36 \\
\rowcolor{gray!6} 
SAT \citep{Springer2021AAttacks}              & 49.74          & 62.45          & $\times$ 4    & $\times$ 8    \\
SWA               & 20.83          & 60.26          & $\times$ 1.00 & $\times$ 1.00 \\
\rowcolor{gray!6} 
l-SAM (ours)            & 48.75          & 85.50          & $\times$ 2    & $\times$ 2    \\
l-AGSAM (ours)           & \textbf{53.14} & \textbf{85.72} & $\times$ 2    & $\times$ 2    \\
\rowcolor{gray!6} 
l-LookSAM (ours)        & 34.92           & 77.49          & $\times$ 1.23 & $\times$ 1.22 \\ \bottomrule
\end{tabular}
\end{center}
\end{table*}

\section{Putting It All Together: Improving Transferability Techniques With Sharpness Minimization}
\label{sec:evalution}

In this section, we show that explicitly minimizing sharpness is a competitive technique for training surrogate models and complements well other transferability techniques. To benchmark our principle against related work, we adhere to the best practices suggested by \cite{Zhao2022TowardsAttacks}. Specifically, we evaluate the benefits of minimizing sharpness on large neighborhoods against other surrogate training techniques (same category), and also assess their complementarity with techniques from distinct categories. All our code and models are available on GitHub\footnote{\tiny{\url{https://github.com/Framartin/rfn-flatness-transferability}}}.

    \paragraph{Minimizing sharpness improves over competitive techniques.}

l-SAM and l-AGSAM are competitive alternatives to existing surrogate training techniques, and l-LookSAM offers good transferability for a small computational overhead. For a fair comparison, we choose the epoch of the early stopped SGD surrogate by evaluating a validation transferability at every training epoch\footnote{To avoid data leakage that violates our no-query threat model, we craft one thousand adversarial examples from images of a validation set and evaluate them on a distinct set of target models.}. We retrieve the SAT (Slight Adversarial Training) ImageNet weights used by \citet{Springer2021AAttacks}, and we train SAT on CIFAR-10 using their hyperparameters. Table~\ref{tab:competitive-ttechs-new} reports the average success rate of the aforementioned techniques, alongside their computational overhead. This overhead is the ratio of forward-backward passes needed by the training technique to those required by SGD. On both datasets, l-AGSAM is the best surrogate. l-AGSAM beats the transferability of SAT, while dividing the training cost by two on ImageNet and four on CIFAR-10. Nevertheless, l-AGSAM doubles the computational number of forward-backward passes compared to SGD. By computing the additional SAM gradient only once per five iterations, l-LookSAM is a viable alternative to contain the computational overhead to 1.23, while having higher transferability than SGD. Overall, sharpness-aware minimizers with large flat neighborhoods offer a good trade-off between transferability and computation.

\begin{table*}[t]
\begin{center}
\caption{Success rate of other categories of transferability techniques applied on the standard SGD base surrogate and on our l-SAM base surrogate. Average success rate on our ten ImageNet targets from ResNet-50 models with a maximum perturbation $L_\infty$ norm $\varepsilon$. Bold is best. In \%.}
\label{tab:complementary-ttechs-new}

\begin{tabular}[t]{lrrlrrl}
\toprule
\multicolumn{1}{c}{ } & \multicolumn{2}{c}{$\varepsilon = 2/255$} & \multicolumn{2}{c}{$\varepsilon = 4/255$} & \multicolumn{2}{c}{$\varepsilon = 8/255$} \\
\cmidrule(l{3pt}r{3pt}){2-3} \cmidrule(l{3pt}r{3pt}){4-5} \cmidrule(l{3pt}r{3pt}){6-7}
Attack & SGD & l-SAM & SGD & l-SAM & SGD & l-SAM\\
\midrule
\addlinespace[0.3em]
\multicolumn{7}{l}{\textbf{Model Augmentation Techniques}}\\
\hspace{1em}\cellcolor{gray!6}{GN \citep{Li2018LearningNetworks}} & \cellcolor{gray!6}{12.9} & \cellcolor{gray!6}{\textbf{28.8}} & \cellcolor{gray!6}{27.8} & \cellcolor{gray!6}{\textbf{52.8}} & \cellcolor{gray!6}{46.5} & \cellcolor{gray!6}{\textbf{71.0}}\\
\hspace{1em}SGM \citep{Wu2020SkipResNets} & 11.7 & \textbf{24.3} & 29.3 & \textbf{51.5} & 55.6 & \textbf{76.2}\\
\hspace{1em}\cellcolor{gray!6}{LGV \citep{Gubri2022LGV:Vicinity}} & \cellcolor{gray!6}{24.8} & \cellcolor{gray!6}{\textbf{25.2}} & \cellcolor{gray!6}{53.5} & \cellcolor{gray!6}{\textbf{54.7}} & \cellcolor{gray!6}{72.1} & \cellcolor{gray!6}{\textbf{73.7}}\\
\addlinespace[0.3em]
\multicolumn{7}{l}{\textbf{Data Augmentation Techniques}}\\
\hspace{1em}DI \citep{Xie2018} & 22.1 & \textbf{42.0} & 47.0 & \textbf{72.5} & 69.4 & \textbf{86.9}\\
\hspace{1em}\cellcolor{gray!6}{SI \citep{Lin2019}} & \cellcolor{gray!6}{10.8} & \cellcolor{gray!6}{\textbf{28.8}} & \cellcolor{gray!6}{26.9} & \cellcolor{gray!6}{\textbf{56.7}} & \cellcolor{gray!6}{49.9} & \cellcolor{gray!6}{\textbf{77.2}}\\
\hspace{1em}VT \citep{Wang2021EnhancingTuning} & 10.5 & \textbf{31.9} & 24.9 & \textbf{59.4} & 43.0 & \textbf{78.5}\\
\addlinespace[0.3em]
\multicolumn{7}{l}{\textbf{Attack Optimizers}}\\
\hspace{1em}\cellcolor{gray!6}{MI \citep{Dong2018BoostingMomentum}} & \cellcolor{gray!6}{12.3} & \cellcolor{gray!6}{\textbf{32.0}} & \cellcolor{gray!6}{26.8} & \cellcolor{gray!6}{\textbf{59.6}} & \cellcolor{gray!6}{46.3} & \cellcolor{gray!6}{\textbf{78.3}}\\
\hspace{1em}NI \citep{Lin2019} & 8.3 & \textbf{20.6} & 22.3 & \textbf{46.5} & 43.9 & \textbf{70.5}\\
\hspace{1em}\cellcolor{gray!6}{RAP \citep{qinBoostingTransferabilityAdversarial2022}} & \cellcolor{gray!6}{11.5} & \cellcolor{gray!6}{\textbf{30.1}} & \cellcolor{gray!6}{26.2} & \cellcolor{gray!6}{\textbf{56.2}} & \cellcolor{gray!6}{42.8} & \cellcolor{gray!6}{\textbf{74.7}} \\
\bottomrule
\end{tabular}
\end{center}
\end{table*}

    \paragraph{Minimizing sharpness trains better base models for complementary techniques.}

l-SAM is a good base model to combine with existing model augmentation, data augmentation, and attack optimization transferability techniques. These categories aim complementary objectives: model and data augmentations reduce the tendency of the attack to overfit the base model by adding randomness to gradients. Attack optimizers intend to smooth the gradient updates. Table~\ref{tab:complementary-ttechs-new} reports the success rate of nine transferability techniques combined with our l-SAM base model on ImageNet. For all perturbation norms $\varepsilon$, l-SAM provides a base model that improves every nine techniques, compared to the standard fully trained SGD surrogate, from 0.4 to 35.5 percentage points. 
RAP \citep{qinBoostingTransferabilityAdversarial2022}) is particularly interesting, since RAP minimizes sharpness like SAM but in the input space. SAM and RAP are best combined, indicating their complementary effects on two distinct factors: SAM finds generic representations, while RAP finds adversarial examples that do not overfit a single representation.

\section{Conclusion}
\label{sec:conclusion}

Overall, our insights into the behavior of SGD through the lens of transferability drive us to a successful approach to train better surrogate models with limited computational overhead. We reject the hypothesis that early stopping benefits transferability due to an inherent trade-off between robust and non-robust features. Instead, we explain the success of early stopping in relation to the dynamics of the exploration of the loss landscape, focusing on sharpness. SGD drives down the valley and progressively falls into deep, sharp holes. These fully trained representations are too specific to generate highly transferable adversarial examples. We remediate this issue by minimizing sharpness. The strong SAM regularization from large flat neighborhoods closely links with transferability. Avoiding large sharp holes proves useful in improving transferability on its own and in complement with existing transferability techniques. Future research could investigate the relationship between transferability and the secondary effects of minimizing sharpness \citep{pmlr-v151-bisla22a,andriushchenkoModernLookRelationship2023,wenSharpnessMinimizationAlgorithms2023}.

\clearpage

\begin{acknowledgements} 
    This work is supported by the Luxembourg National Research Funds (FNR) through CORE project C18/IS/12669767/STELLAR/LeTraon. This work was also supported by NAVER Corporation.

\end{acknowledgements}

\bibliography{references}

\begin{thebibliography}{37}
\providecommand{\natexlab}[1]{#1}
\providecommand{\url}[1]{\texttt{#1}}
\expandafter\ifx\csname urlstyle\endcsname\relax
  \providecommand{\doi}[1]{doi: #1}\else
  \providecommand{\doi}{doi: \begingroup \urlstyle{rm}\Url}\fi

\bibitem[Andriushchenko et~al.(2023)Andriushchenko, Croce, M{\"u}ller, Hein, and Flammarion]{andriushchenkoModernLookRelationship2023}
Maksym Andriushchenko, Francesco Croce, Maximilian M{\"u}ller, Matthias Hein, and Nicolas Flammarion.
\newblock A modern look at the relationship between sharpness and generalization.
\newblock February 2023.

\bibitem[Ashukha et~al.(2020)Ashukha, Lyzhov, Molchanov, and Vetrov]{Ashukha2020PitfallsLearning}
Arsenii Ashukha, Alexander Lyzhov, Dmitry Molchanov, and Dmitry Vetrov.
\newblock {Pitfalls of In-Domain Uncertainty Estimation and Ensembling in Deep Learning}.
\newblock 2 2020.
\newblock URL \url{http://arxiv.org/abs/2002.06470}.

\bibitem[Benz et~al.(2021)Benz, Zhang, and Kweon]{Benz2021BatchPerspective}
Philipp Benz, Chaoning Zhang, and In~So Kweon.
\newblock {Batch Normalization Increases Adversarial Vulnerability and Decreases Adversarial Transferability: A Non-Robust Feature Perspective}.
\newblock In \emph{ICCV 2021}, 10 2021.
\newblock \doi{10.1109/ICCV48922.2021.00772}.
\newblock URL \url{http://arxiv.org/abs/2010.03316}.

\bibitem[Biggio et~al.(2013)Biggio, Corona, Maiorca, Nelson, {\v{S}}rndi{\'{c}}, Laskov, Giacinto, and Roli]{Biggio2013}
Battista Biggio, Igino Corona, Davide Maiorca, Blaine Nelson, Nedim {\v{S}}rndi{\'{c}}, Pavel Laskov, Giorgio Giacinto, and Fabio Roli.
\newblock {Evasion attacks against machine learning at test time}.
\newblock In \emph{Lecture Notes in Computer Science (including subseries Lecture Notes in Artificial Intelligence and Lecture Notes in Bioinformatics)}, volume 8190 LNAI, pages 387--402, 8 2013.
\newblock ISBN 9783642409936.
\newblock \doi{10.1007/978-3-642-40994-3{\_}25}.
\newblock URL \url{http://arxiv.org/abs/1708.06131 http://dx.doi.org/10.1007/978-3-642-40994-3_25}.

\bibitem[Bisla et~al.(2022)Bisla, Wang, and Choromanska]{pmlr-v151-bisla22a}
Devansh Bisla, Jing Wang, and Anna Choromanska.
\newblock Low-pass filtering sgd for recovering flat optima in the deep learning optimization landscape.
\newblock In Gustau Camps-Valls, Francisco J.~R. Ruiz, and Isabel Valera, editors, \emph{Proceedings of The 25th International Conference on Artificial Intelligence and Statistics}, volume 151 of \emph{Proceedings of Machine Learning Research}, pages 8299--8339. PMLR, 28--30 Mar 2022.
\newblock URL \url{https://proceedings.mlr.press/v151/bisla22a.html}.

\bibitem[Dong et~al.(2018)Dong, Liao, Pang, Su, Zhu, Hu, and Li]{Dong2018BoostingMomentum}
Yinpeng Dong, Fangzhou Liao, Tianyu Pang, Hang Su, Jun Zhu, Xiaolin Hu, and Jianguo Li.
\newblock {Boosting Adversarial Attacks with Momentum}.
\newblock In \emph{Proceedings of the IEEE Computer Society Conference on Computer Vision and Pattern Recognition}, pages 9185--9193, 10 2018.
\newblock ISBN 9781538664209.
\newblock \doi{10.1109/CVPR.2018.00957}.
\newblock URL \url{http://arxiv.org/abs/1710.06081}.

\bibitem[Du et~al.(2021)Du, Yan, Feng, Zhou, Zhen, Goh, and Tan]{duEfficientSharpnessawareMinimization2021}
Jiawei Du, Hanshu Yan, Jiashi Feng, Joey~Tianyi Zhou, Liangli Zhen, Rick Siow~Mong Goh, and Vincent Tan.
\newblock Efficient {{Sharpness-aware Minimization}} for {{Improved Training}} of {{Neural Networks}}.
\newblock In \emph{International {{Conference}} on {{Learning Representations}}}, October 2021.

\bibitem[Engstrom et~al.(2019)Engstrom, Ilyas, Santurkar, and Tsipras]{Engstrom2019RobustnessLibrary}
Logan Engstrom, Andrew Ilyas, Shibani Santurkar, and Dimitris Tsipras.
\newblock {Robustness (Python Library)}, 2019.
\newblock URL \url{https://github.com/MadryLab/robustness}.

\bibitem[Foret et~al.(2020)Foret, Kleiner Google~Research, Mobahi Google~Research, and Neyshabur~Blueshift]{Foret2020Sharpness-AwareGeneralization}
Pierre Foret, Ariel Kleiner Google~Research, Hossein Mobahi Google~Research, and Behnam Neyshabur~Blueshift.
\newblock {Sharpness-Aware Minimization for Efficiently Improving Generalization}.
\newblock 10 2020.
\newblock \doi{10.48550/arxiv.2010.01412}.
\newblock URL \url{https://arxiv.org/abs/2010.01412v3}.

\bibitem[Goodfellow et~al.(2014)Goodfellow, Shlens, and Szegedy]{Goodfellow2014ExplainingExamples}
Ian~J. Goodfellow, Jonathon Shlens, and Christian Szegedy.
\newblock {Explaining and Harnessing Adversarial Examples}.
\newblock 12 2014.
\newblock URL \url{http://arxiv.org/abs/1412.6572}.

\bibitem[Gubri et~al.(2022{\natexlab{a}})Gubri, Cordy, Papadakis, Le~Traon, and Sen]{Gubri2022EfficientNetworks}
Martin Gubri, Maxime Cordy, Mike Papadakis, Yves Le~Traon, and Koushik Sen.
\newblock {Efficient and Transferable Adversarial Examples from Bayesian Neural Networks}.
\newblock In \emph{UAI 2022}, 2022{\natexlab{a}}.
\newblock URL \url{http://arxiv.org/abs/2011.05074}.

\bibitem[Gubri et~al.(2022{\natexlab{b}})Gubri, Cordy, Papadakis, Traon, and Sen]{Gubri2022LGV:Vicinity}
Martin Gubri, Maxime Cordy, Mike Papadakis, Yves~Le Traon, and Koushik Sen.
\newblock {LGV: Boosting Adversarial Example Transferability from Large Geometric Vicinity}.
\newblock In \emph{ECCV 2022}, 2022{\natexlab{b}}.

\bibitem[Ilyas et~al.(2019)Ilyas, Santurkar, Tsipras, Engstrom, Tran, and Madry]{Ilyas2019AdversarialFeatures}
Andrew Ilyas, Shibani Santurkar, Dimitris Tsipras, Logan Engstrom, Brandon Tran, and Aleksander Madry.
\newblock {Adversarial Examples Are Not Bugs, They Are Features}.
\newblock 5 2019.
\newblock URL \url{http://arxiv.org/abs/1905.02175}.

\bibitem[Izmailov et~al.(2018)Izmailov, Podoprikhin, Garipov, Vetrov, and Wilson]{izmailovAveragingWeightsLeads2018}
Pavel Izmailov, Dmitrii Podoprikhin, Timur Garipov, Dmitry Vetrov, and Andrew~Gordon Wilson.
\newblock Averaging weights leads to wider optima and better generalization.
\newblock In \emph{34th {{Conference}} on {{Uncertainty}} in {{Artificial Intelligence}} 2018, {{UAI}} 2018}, volume~2, pages 876--885. {Association For Uncertainty in Artificial Intelligence (AUAI)}, March 2018.
\newblock ISBN 978-1-5108-7160-1.

\bibitem[Kaddour et~al.(2022)Kaddour, Liu, Silva, and Kusner]{Kaddour2022WhenWork}
Jean Kaddour, Linqing Liu, Ricardo Silva, and Matt~J. Kusner.
\newblock {When Do Flat Minima Optimizers Work?}
\newblock In \emph{NeurIPS 2022}, 2 2022.
\newblock \doi{10.48550/arxiv.2202.00661}.
\newblock URL \url{https://arxiv.org/abs/2202.00661v5}.

\bibitem[Keskar et~al.(2017)Keskar, Mudigere, Nocedal, Smelyanskiy, and Tang]{keskar2017on}
Nitish~Shirish Keskar, Dheevatsa Mudigere, Jorge Nocedal, Mikhail Smelyanskiy, and Ping Tak~Peter Tang.
\newblock On large-batch training for deep learning: Generalization gap and sharp minima.
\newblock In \emph{International Conference on Learning Representations}, 2017.
\newblock URL \url{https://openreview.net/forum?id=H1oyRlYgg}.

\bibitem[Kim(2020)]{Kim2020Torchattacks:Attacks}
Hoki Kim.
\newblock {Torchattacks: A pytorch repository for adversarial attacks}.
\newblock \emph{arXiv preprint arXiv:2010.01950}, 2020.

\bibitem[Kurakin et~al.(2017)Kurakin, Goodfellow, and Bengio]{Kurakin2017}
Alexey Kurakin, Ian~J. Goodfellow, and Samy Bengio.
\newblock {Adversarial examples in the physical world}.
\newblock In \emph{5th International Conference on Learning Representations, ICLR 2017 - Workshop Track Proceedings}, 7 2017.
\newblock URL \url{http://arxiv.org/abs/1607.02533}.

\bibitem[Kwon et~al.(2021)Kwon, Kim, Park, and Choi]{Kwon2021ASAM:Networks}
Jungmin Kwon, Jeongseop Kim, Hyunseo Park, and In~Kwon Choi.
\newblock {ASAM: Adaptive Sharpness-Aware Minimization for Scale-Invariant Learning of Deep Neural Networks}.
\newblock 2 2021.
\newblock \doi{10.48550/arxiv.2102.11600}.
\newblock URL \url{https://arxiv.org/abs/2102.11600v3}.

\bibitem[Li et~al.(2018)Li, Bai, Zhou, Xie, Zhang, and Yuille]{Li2018LearningNetworks}
Yingwei Li, Song Bai, Yuyin Zhou, Cihang Xie, Zhishuai Zhang, and Alan Yuille.
\newblock {Learning Transferable Adversarial Examples via Ghost Networks}.
\newblock \emph{Proceedings of the AAAI Conference on Artificial Intelligence}, 34\penalty0 (07):\penalty0 11458--11465, 12 2018.
\newblock ISSN 2374-3468.
\newblock \doi{10.1609/aaai.v34i07.6810}.
\newblock URL \url{http://arxiv.org/abs/1812.03413}.

\bibitem[Lin et~al.(2019)Lin, Song, He, Wang, and Hopcroft]{Lin2019}
Jiadong Lin, Chuanbiao Song, Kun He, Liwei Wang, and John~E. Hopcroft.
\newblock {Nesterov Accelerated Gradient and Scale Invariance for Adversarial Attacks}.
\newblock 8 2019.
\newblock URL \url{http://arxiv.org/abs/1908.06281}.

\bibitem[Liu et~al.(2022)Liu, Mai, Chen, Hsieh, and You]{liuEfficientScalableSharpnessAware2022}
Yong Liu, Siqi Mai, Xiangning Chen, Cho-Jui Hsieh, and Yang You.
\newblock Towards {{Efficient}} and {{Scalable Sharpness-Aware Minimization}}.
\newblock In \emph{{{CVPR}}}, pages 12360--12370, 2022.

\bibitem[Naseer et~al.(2022)Naseer, Ranasinghe, Khan, Shahbaz~Khan, and Porikli]{Naseer2022OnTransformers}
Muzammal Naseer, Kanchana Ranasinghe, Salman Khan, Fahad Shahbaz~Khan, and Fatih Porikli.
\newblock {On Improving Adversarial Transferability of Vision Transformers}.
\newblock In \emph{ICLR (spotlight)}, 3 2022.

\bibitem[Nitin(2021)]{Nitin2021SGDNon-Robust}
Vikram Nitin.
\newblock {SGD on Neural Networks learns Robust Features before Non-Robust}, 3 2021.

\bibitem[Qin et~al.(2022)Qin, Fan, Liu, Shen, Zhang, Wang, and Wu]{qinBoostingTransferabilityAdversarial2022}
Zeyu Qin, Yanbo Fan, Yi~Liu, Li~Shen, Yong Zhang, Jue Wang, and Baoyuan Wu.
\newblock Boosting the {{Transferability}} of {{Adversarial Attacks}} with {{Reverse Adversarial Perturbation}}.
\newblock In \emph{{{NeurIPS}} 2022}, October 2022.
\newblock \doi{10.48550/arxiv.2210.05968}.

\bibitem[Schneider et~al.(2021)Schneider, Dangel, and Hennig]{Schneider2021Cockpit:Networks}
Frank Schneider, Felix Dangel, and Philipp Hennig.
\newblock {Cockpit: A Practical Debugging Tool for the Training of Deep Neural Networks}.
\newblock 2 2021.
\newblock \doi{10.48550/arxiv.2102.06604}.
\newblock URL \url{https://arxiv.org/abs/2102.06604v2 http://arxiv.org/abs/2102.06604}.

\bibitem[Springer et~al.(2021)Springer, Mitchell, and Kenyon]{Springer2021AAttacks}
Jacob~M. Springer, Melanie Mitchell, and Garrett~T. Kenyon.
\newblock {A Little Robustness Goes a Long Way: Leveraging Robust Features for Targeted Transfer Attacks}.
\newblock \emph{Advances in Neural Information Processing Systems}, 12:\penalty0 9759--9773, 6 2021.
\newblock ISSN 10495258.
\newblock \doi{10.48550/arxiv.2106.02105}.
\newblock URL \url{https://arxiv.org/abs/2106.02105v2}.

\bibitem[Stutz et~al.(2021)Stutz, Hein, and Schiele]{Stutz2021RelatingMinima}
David Stutz, Matthias Hein, and Bernt Schiele.
\newblock {Relating Adversarially Robust Generalization to Flat Minima}.
\newblock \emph{Proceedings of the IEEE International Conference on Computer Vision}, pages 7787--7797, 4 2021.
\newblock ISSN 15505499.
\newblock \doi{10.1109/ICCV48922.2021.00771}.
\newblock URL \url{https://arxiv.org/abs/2104.04448v2}.

\bibitem[Szegedy et~al.(2013)Szegedy, Zaremba, Sutskever, Bruna, Erhan, Goodfellow, and Fergus]{Szegedy2013}
Christian Szegedy, Wojciech Zaremba, Ilya Sutskever, Joan Bruna, Dumitru Erhan, Ian Goodfellow, and Rob Fergus.
\newblock {Intriguing properties of neural networks}.
\newblock 12 2013.
\newblock URL \url{http://arxiv.org/abs/1312.6199}.

\bibitem[Wang and He(2021)]{Wang2021EnhancingTuning}
Xiaosen Wang and Kun He.
\newblock {Enhancing the Transferability of Adversarial Attacks through Variance Tuning}.
\newblock \emph{Proceedings of the IEEE Computer Society Conference on Computer Vision and Pattern Recognition}, pages 1924--1933, 3 2021.
\newblock ISSN 10636919.
\newblock \doi{10.1109/CVPR46437.2021.00196}.
\newblock URL \url{https://arxiv.org/abs/2103.15571v3}.

\bibitem[Wen et~al.(2023)Wen, Li, and Ma]{wenSharpnessMinimizationAlgorithms2023}
Kaiyue Wen, Zhiyuan Li, and Tengyu Ma.
\newblock Sharpness {{Minimization Algorithms Do Not Only Minimize Sharpness To Achieve Better Generalization}}, July 2023.

\bibitem[Wu et~al.(2020)Wu, Wang, Xia, Bailey, and Ma]{Wu2020SkipResNets}
Dongxian Wu, Yisen Wang, Shu-Tao Xia, James Bailey, and Xingjun Ma.
\newblock {Skip Connections Matter: On the Transferability of Adversarial Examples Generated with ResNets}.
\newblock In \emph{ICLR}, 2 2020.
\newblock URL \url{https://arxiv.org/abs/2002.05990v1 http://arxiv.org/abs/2002.05990}.

\bibitem[Xie et~al.(2019)Xie, Zhang, Zhou, Bai, Wang, Ren, and Yuille]{Xie2018}
Cihang Xie, Zhishuai Zhang, Yuyin Zhou, Song Bai, Jianyu Wang, Zhou Ren, and Alan~L. Yuille.
\newblock {Improving transferability of adversarial examples with input diversity}.
\newblock In \emph{Proceedings of the IEEE Computer Society Conference on Computer Vision and Pattern Recognition}, volume 2019-June, pages 2725--2734, 3 2019.
\newblock ISBN 9781728132938.
\newblock \doi{10.1109/CVPR.2019.00284}.
\newblock URL \url{http://arxiv.org/abs/1803.06978}.

\bibitem[Yao et~al.(2019)Yao, Gholami, Keutzer, and Mahoney]{Yao2019PyHessian:Hessian}
Zhewei Yao, Amir Gholami, Kurt Keutzer, and Michael~W. Mahoney.
\newblock {PyHessian: Neural Networks Through the Lens of the Hessian}.
\newblock \emph{Proceedings - 2020 IEEE International Conference on Big Data, Big Data 2020}, pages 581--590, 12 2019.
\newblock ISSN 2331-8422.
\newblock \doi{10.1109/BigData50022.2020.9378171}.
\newblock URL \url{https://arxiv.org/abs/1912.07145v3}.

\bibitem[Zhang et~al.(2021)Zhang, Cho, Benz, Zhang, Zhang, Youn, and Kweon]{Zhang2021EarlyTransferability}
Chaoning Zhang, Gyusang Cho, Philipp Benz, Kang Zhang, Chenshuang Zhang, Chan-Hyun Youn, and In~So Kweon.
\newblock {Early Stop And Adversarial Training Yield Better surrogate Model: Very Non-Robust Features Harm Adversarial Transferability}, 2021.

\bibitem[Zhao et~al.(2022)Zhao, Zhang, Li, Sicre, Amsaleg, and Backes]{Zhao2022TowardsAttacks}
Zhengyu Zhao, Hanwei Zhang, Renjue Li, Ronan Sicre, Laurent Amsaleg, and Michael Backes.
\newblock {Towards Good Practices in Evaluating Transfer Adversarial Attacks}.
\newblock 11 2022.
\newblock \doi{10.48550/arxiv.2211.09565}.
\newblock URL \url{https://arxiv.org/abs/2211.09565v1}.

\bibitem[Zhuang et~al.(2022)Zhuang, Gong, Yuan, Cui, Adam, Dvornek, {sekhar tatikonda}, s~Duncan, and Liu]{Zhuang2022SurrogateTraining}
Juntang Zhuang, Boqing Gong, Liangzhe Yuan, Yin Cui, Hartwig Adam, Nicha~C Dvornek, {sekhar tatikonda}, James s~Duncan, and Ting Liu.
\newblock {Surrogate Gap Minimization Improves Sharpness-Aware Training}.
\newblock In \emph{International Conference on Learning Representations}, 2022.
\newblock URL \url{https://openreview.net/forum?id=edONMAnhLu-}.

\end{thebibliography}


\newpage

\onecolumn

\title{Going Further: Flatness at the Rescue of Early Stopping for Adversarial Example Transferability\\(Supplementary Material)}
\maketitle

\appendix

These supplementary materials contain the following sections:

\begin{itemize}
    \item Appendix \ref{sec:app-xp-settings} details the experimental settings,
    \item Appendix \ref{sec:app-transferability-epochs} reports the transferability and the natural accuracy by epochs of surrogate trained with SGD on CIFAR-10 and ImageNet,
    \item Appendix \ref{sec:app-rfs-nrfs} reports additional results of Section \ref{sec:early-stopping} ``Another Look at the Non-Robust Features Hypothesis about Early Stopping'',
    \item Appendix \ref{sec:app-training-dynamics} reports additional results of Section \ref{sec:training-dynamics} ``Stopping Earlier: Transferability and Training Dynamics'',
    \item Appendix \ref{sec:app-sam-hp} reports additional results of Section \ref{sec:sam} ``Going Further: Flatness At The Rescue of SGD'',
    \item Appendix \ref{sec:app-wd} reports the results about transferability with respect to the weight decay of the surrogate,
    \item Appendix \ref{sec:app-eval-rfn} reports additional results of Section \ref{sec:evalution} ``Putting It All Together: Improving Transferability Techniques With Sharpness Minimization''.
\end{itemize}

\section{Experimental Settings}
\label{sec:app-xp-settings}

This section describes the experimental settings used in this article. The experimental setup is standard for transfer-based attacks.

\begin{itemize}
    \item Our source code used to train and evaluate models is publicly available on GitHub at this URL: \newline \url{https://github.com/Framartin/rfn-flatness-transferability}.
    \item  Our trained models on both CIFAR-10 and ImageNet are publicly distributed through HuggingFace at this URL: \url{https://huggingface.co/mgubri/rfn-flatness-transferability}.
\end{itemize}

\paragraph{Target models.} All our target models on CIFAR-10 are fully trained for 150 epochs with SGD using the hyperparameters reported in Table \ref{tab:hp-training}. For a fair comparison, \textit{the baseline surrogate is trained with SGD using the same hyperparameters as the targets}. On CIFAR-10, we target the following ten architectures: ResNet-50 (the surrogate with the same architecture is an independently trained model), ResNet-18, ResNet-101, DenseNet-161, DenseNet-201, WideResNet-28-10, VGG13, VGG19, Inception v3 and ConvMixer. On ImageNet, the target models are the pretrained models distributed by PyTorch. The ten target architectures on ImageNet are the following: ResNet-50, ResNet-152, ResNeXt-50 32X4D, WideResNet-50-2, DenseNet-201, VGG19, GoogLeNet (Inception v1), Inception v3, ViT B 16 and Swin S. Additionally, we train a ``validation'' set of architectures on CIFAR-10 to select hyperparameters independently of reported results. This set is composed of: ResNet-50 (another independently trained model), ResNet-34, ResNet-152, DenseNet-121, DenseNet-169, WideResNet-16-8, VGG11, VGG16, GoogLeNet (Inception v1) and MLPMixer. This validation set of target models on ImageNet is composed of the following architectures: ResNet-50 (another independently trained model), ResNet-101, ResNeXt-101 64X4D, WideResNet101-2, VGG16, DenseNet121, ViT B 32 and Swin B.

\paragraph{Surrogate models trained with SGD.} We train the surrogate models on CIFAR-10 and ImageNet using SGD with the standard hyperparameters of the robustness library~\citep{Engstrom2019RobustnessLibrary} (Table \ref{tab:hp-training}). Due to computational limitations on ImageNet, we limit the number of epochs to 90, reusing the same hyperparameters as \citet{Ashukha2020PitfallsLearning}.

\paragraph{Surrogate models trained with SAM and its variants.} We train surrogate models with SAM using the same hyperparameters as the models trained with SGD for both datasets. We integrate the SAM optimizer into the robustness library~\cite{Engstrom2019RobustnessLibrary}. The unique hyperparameter of SAM is $\rho$, which is set to $0.05$ as the original paper for both datasets for the original SAM surrogate. Our SAM surrogate with large flat neighborhoods, called l-SAM, is trained with SAM with $\rho$ equal to $0.4$. The $\rho$ values used to train the variants of SAM are reported in Table \ref{tab:hp-training}. We use official or popular implementations of ASAM \citep{Kwon2021ASAM:Networks}, GSAM \citep{Zhuang2022SurrogateTraining}, AGSAM (GSAM+ASAM), LookSAM \citep{liuEfficientScalableSharpnessAware2022} and WASAM (SAM+SWA, \citet{Kaddour2022WhenWork}), following the original papers. 
LookSAM is an efficient variant of SAM that computes the additional gradient of SAM only once per five training iterations.  As reported by \citet{liuEfficientScalableSharpnessAware2022}, LookSAM is unstable at the beginning of training. \citet{liuEfficientScalableSharpnessAware2022} solve this issue using a learning rate with warmup. Since we wanted to use the same learning rate schedule for all training techniques, we added another type of warmup. LookSAM computes the additional SAM gradient at all training iterations during the first three epochs. Our LookSAM is equivalent to SAM before the fourth epoch. This simple solution is enough for LookSAM to converge. This computational overhead is taken into account in the computational cost reported in Table \ref{tab:competitive-ttechs-new}.

\paragraph{Other surrogate training techniques (Section 6).} To compare with competitive training techniques on ImageNet, we retrieve the original models of SAT \cite{Springer2021AAttacks}, an adversarially trained model with a small maximum $L_2$ norm perturbation $\varepsilon$ of $0.1$ and with the PGD attack applied with 3 steps and a step size equal to $2 \varepsilon / 3$. On CIFAR-10, we reuse the best hyperparameters of \cite{Springer2021AAttacks} to adversarially train the SAT surrogate model with a maximum $L_2$ norm $\varepsilon$ of $0.025$ and PGD with 7 steps and a step size of $0.3 \varepsilon$. For a fair comparison, we choose the best checkpoint of the early stopped SGD surrogate by evaluating the transferability of every training epoch. For each epoch, we craft 1,000 adversarial examples from a distinct validation set of original examples and compute their success rate over a distinct set of validation target architectures. On CIFAR-10, the selected epoch is 54, and 66 on ImageNet. All the other hyperparameters not mentioned in this paragraph are the same as those used to train the surrogates with SGD.

\paragraph{Non-surrogate training transferability techniques (Section 6).} We consider nine transferability techniques that are not surrogate training techniques, i.e., model augmentation (GN, SGM, LGV), data augmentation (DI, SI, VT), attack optimizers (MI, NI, RAP). We study their complementarity with our surrogate training technique. SI uses $m=5$ copies. VT uses $\beta=1.8$. GN uses a random range of $[1-0.3,1+0.3]$. We use the following hyperparameters. MI uses a decay of 1.2 and NI a decay of 0.6. DI uses a resize rate of 0.85 with a diversity probability of 0.8. We apply the inner optimization of RAP (gradient ascent) with 5 steps and a search neighborhood $\varepsilon_n = \frac{2}{3}\varepsilon$ to keep the original proportion. Our preliminary experiments confirm the observation of \citet{qinBoostingTransferabilityAdversarial2022} that starting the inner optimization from the first outer optimization step is not optimal. For fairness to RAP, we start the inner optimization at the 10th step of the outer optimization.

\paragraph{Attack.} Unless specified otherwise, we use the BIM (Basic Iterative Method, equivalently called I-FGSM) \cite{Kurakin2017} which is the standard attack for transferability \cite{Benz2021BatchPerspective,Dong2018BoostingMomentum,Gubri2022EfficientNetworks,Gubri2022LGV:Vicinity,Li2018LearningNetworks,Lin2019,Springer2021AAttacks,Wu2020SkipResNets,Xie2018,Zhao2022TowardsAttacks,qinBoostingTransferabilityAdversarial2022}. By default, the maximum $L_\infty$ perturbation norm $\varepsilon$ is set to $4/255$. We use the BIM hyperparameters tuned by \cite{Gubri2022EfficientNetworks,Gubri2022LGV:Vicinity} on a distinct set of validation target models: BIM performs 50 iterations with a step size equal to $\varepsilon/10$. Unless specified otherwise, we craft adversarial examples from a subset of 1,000 natural test examples that are correctly predicted by all target models. We repeat the experiments on CIFAR-10 three times, each run with a different random seed, an independently sampled subset of original examples, and \textit{an independently trained surrogate model}. For every CIFAR-10 experiment, we train three times each surrogate model to estimate correctly the randomness of an attacker training a surrogate model to perform an attack. The success rate is the misclassification rate of these adversarial examples evaluated on one target model. We report the average success rate across the three random seeds, along with a confidence interval of plus/minus two times the empirical standard deviation. 

\paragraph{Threat model.} We study the threat model of untargeted adversarial examples: the adversary's goal is misclassification. We consider the standard adversary capability for transfer-based black-box attacks, where the adversary does not have query access to the target model. Query-based attacks are another distinct family of attacks. The attacker does not know the weights of the target model, nor its architecture. Similarly to all the related work \citep{Benz2021BatchPerspective,Dong2018BoostingMomentum,Gubri2022EfficientNetworks,Gubri2022LGV:Vicinity,Li2018LearningNetworks,Lin2019,Springer2021AAttacks,Wu2020SkipResNets,Xie2018,Zhao2022TowardsAttacks,qinBoostingTransferabilityAdversarial2022}, we suppose that the attacker can independently train a surrogate model on the same training data.

\paragraph{Sharpness metrics.} We compute both sharpness metrics (Hessian top-1 eigenvalue and Hessian trace) at every epoch using the PyHessian library \citep{Yao2019PyHessian:Hessian} on a random subset of a thousand examples from the CIFAR-10 train dataset.

\paragraph{Implementation.} The source code for each experiment is available on GitHub. Our models are distributed through HuggingFace. We use the torchattacks library \citep{Kim2020Torchattacks:Attacks} to craft adversarial examples with the BIM attacks and four transferability techniques, namely LGV, DI, SI, VT, MI and NI. We reuse the original implementations of GN and SGM to ``patch'' the surrogate architecture, and use the TorchAttacks implementation of BIM on top. We adapted the original implementation of RAP to fit our experimental settings. The software versions are the following: Python 3.10.8, PyTorch 1.12.1, Torchvision 0.13.1, and TorchAttacks 3.3.0.

\paragraph{Infrastructure.} For all experiments, we use Tesla V100-DGXS-32GB GPUs on a server with 256GB of RAM, CUDA 11.4, and the Ubuntu operating system.

\begin{table}[h]
\begin{center}
\caption{Hyperparameters used to train surrogate models.}
\label{tab:hp-training}
\begin{tabular}{@{}llll@{}}
\toprule
\multicolumn{1}{c}{\textbf{Training}} & \multicolumn{1}{c}{\textbf{Hyperparameter}} & \multicolumn{1}{c}{\textbf{Dataset}} & \textbf{Value} \\ \midrule
\multirow{10}{*}{All} & \multirow{2}{*}{Number of epochs}    & CIFAR-10 & 150                          \\ \cmidrule(l){3-4} 
                      &                                      & ImageNet & 90                           \\ \cmidrule(l){2-4} 
                      & Initial learning rate                & All      & 0.1                          \\ \cmidrule(l){2-4} 
                      & \multirow{2}{*}{Learning rate decay} & CIFAR-10 & Step-wise /10 each 50 epochs \\ \cmidrule(l){3-4} 
                      &                                      & ImageNet & Step-wise /10 each 30 epochs \\ \cmidrule(l){2-4} 
                      & Momentum                             & All      & 0.9                          \\ \cmidrule(l){2-4} 
                      & \multirow{2}{*}{Batch-size}          & CIFAR-10 & 128                          \\ \cmidrule(l){3-4} 
                      &                                      & ImageNet & 256                          \\ \cmidrule(l){2-4} 
                      & \multirow{2}{*}{Weight decay}        & CIFAR-10 & 0.0005                       \\ \cmidrule(l){3-4} 
                      &                                      & ImageNet & 0.0001                       \\ \midrule
SAM                   & $\rho$                               & All      & 0.05 for SAM, 0.4 for l-SAM    \\ \midrule
\multirow{2}{*}{GSAM} & $\rho$                               & All      & 0.05 for GSAM, 0.2 for l-GSAM    \\ 
                      & $\alpha$                             & All      & 0.15    \\ \midrule
\multirow{3}{*}{LookSAM} & $\rho$                            & All      & 0.05 for LookSAM, 0.3 for l-LookSAM    \\ 
                      & $k$                                  & All      & 5     \\
                      & SAM Warmup                           & All      & 3 epochs     \\ \midrule
ASAM                  & $\rho$                               & All      & 0.5 for ASAM, 3 for l-ASAM    \\  \midrule
\multirow{2}{*}{AGSAM} & $\rho$                              & All      & 0.5 for AGSAM, 4 for l-AGSAM    \\ 
                      & $\alpha$                             & All      & 0.15    \\ 
\bottomrule
\end{tabular}
\end{center}
\end{table}

\clearpage
\section{Transferability and Natural Accuracy by Epochs}
\label{sec:app-transferability-epochs}

\begin{figure}[h]
\begin{center}
   \includegraphics[width=0.5\linewidth]{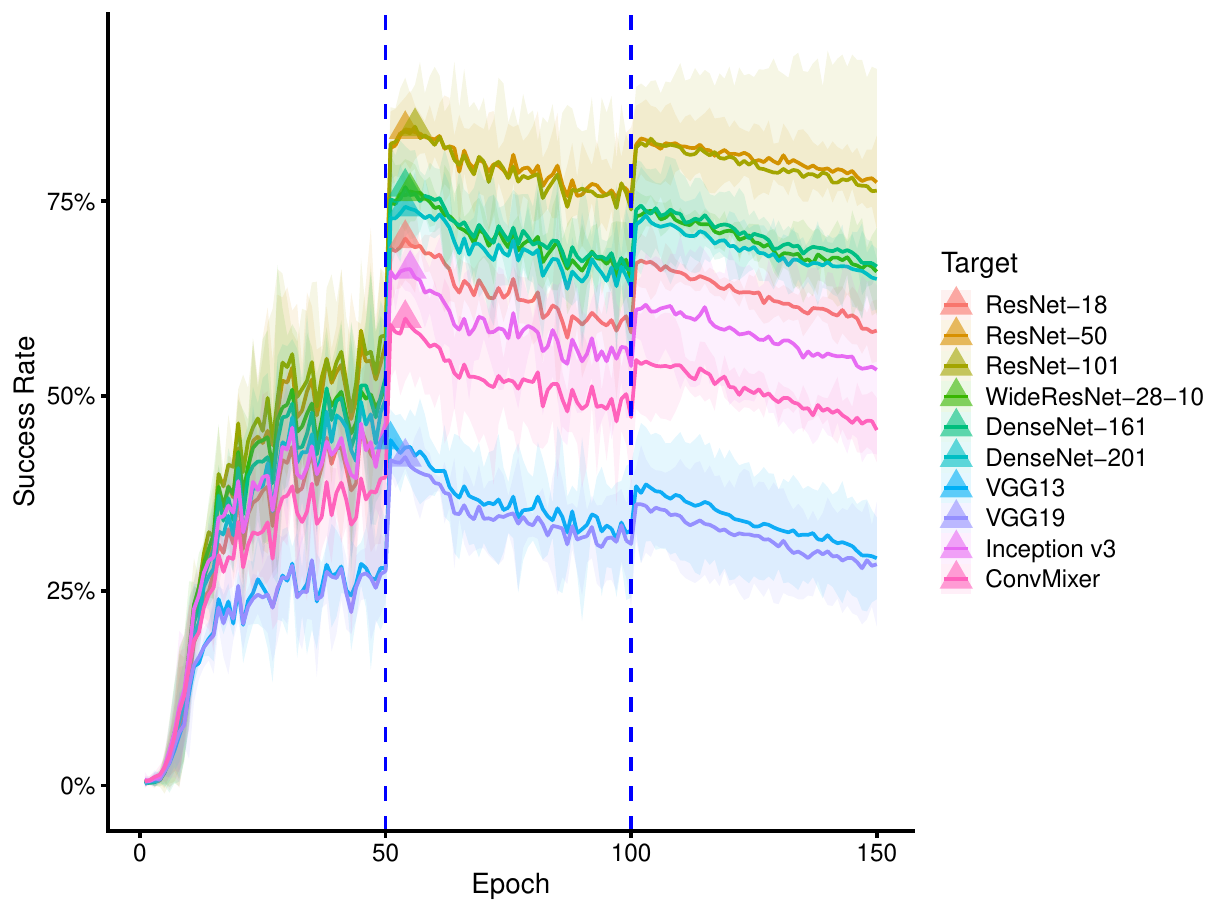}
\end{center}
   \caption{Early stopping improves transferability consistently across target models on CIFAR-10. Success rate evaluated on ten target models (color) from a ResNet-50 surrogate model trained for a number of epochs (x-axis) on the CIFAR-10 dataset. We report the average over three random seeds (line) and the confidence interval of two standard deviations (colored area). Vertical bars indicate the step decays of the learning rate. Triangles indicate the epochs corresponding to the highest success rate per target.}
\label{fig:plot_successrate_epochs_standard_schedule}
\end{figure}

\begin{figure}[h]
\begin{center}
   \includegraphics[width=0.7\linewidth]{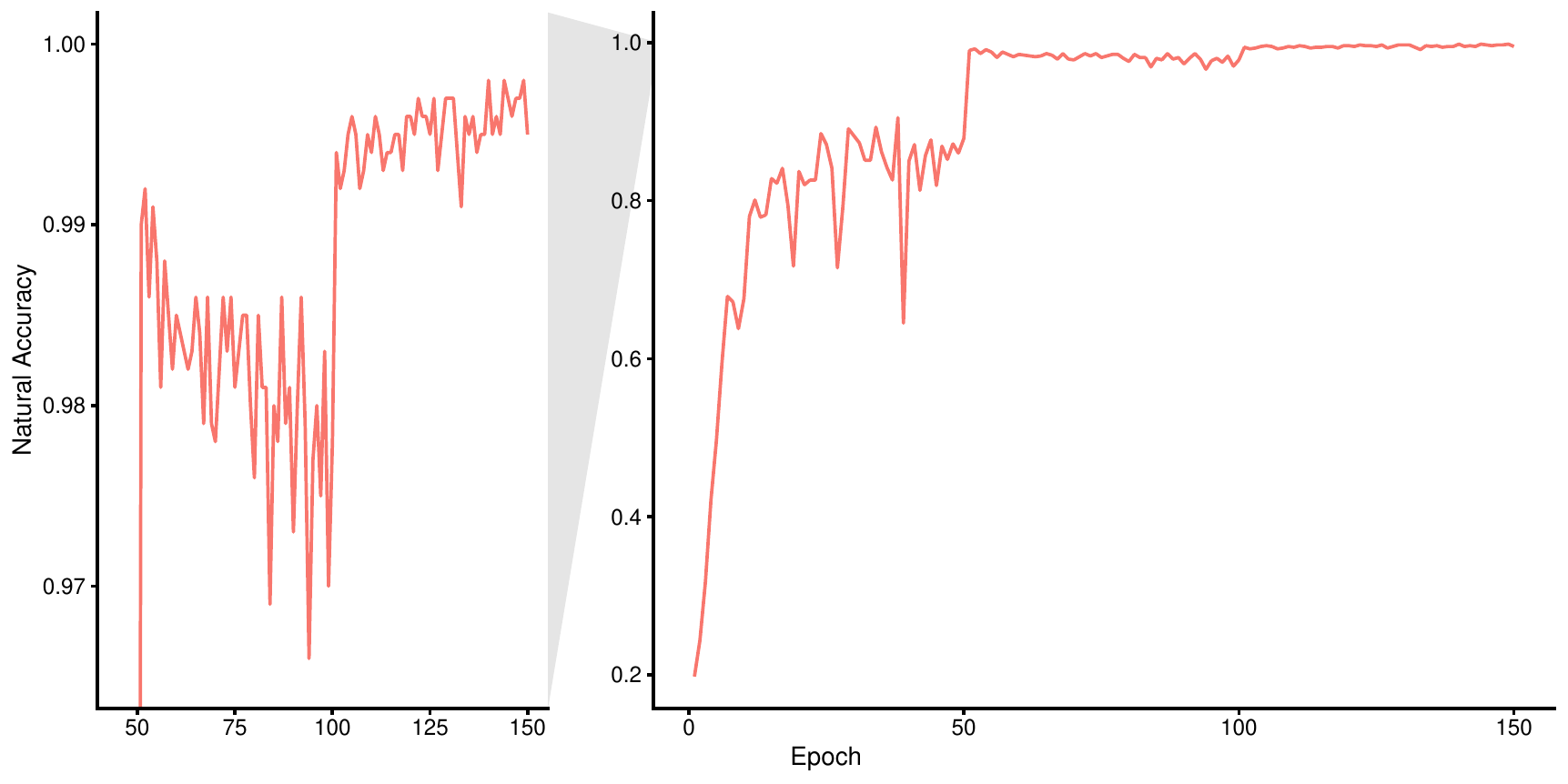}
\end{center}
   \caption{The natural accuracy increases at the end of training, whereas transferability decreases. Natural test accuracy of the ResNet-50 surrogate model trained for a number of epochs (x-axis) on the CIFAR-10 dataset. Evaluated on the test subset used to craft adversarial examples in Figure \ref{fig:plot_successrate_epochs_standard_schedule}.}
\label{fig:plot_acc_epochs_standard}
\end{figure}

\begin{figure}[h]
\begin{center}
   \includegraphics[width=0.5\linewidth]{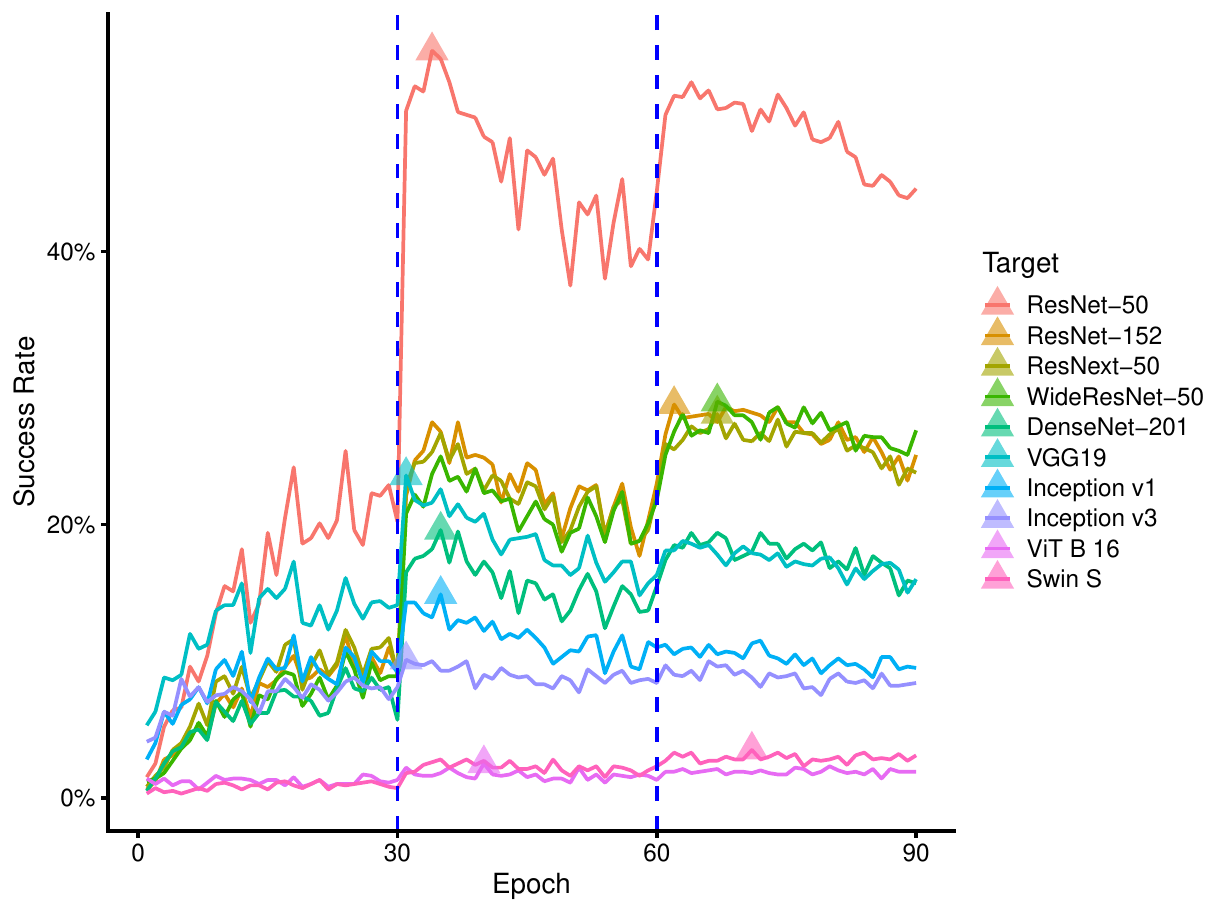}
\end{center}
   \caption{Early stopping improves transferability to various target models on ImageNet, except to vision transformers (ViT-B-16 and Swin-S) against which the success rate plateaus at the end of training. Success rate evaluated on ten target models (colour) from a ResNet-50 surrogate model trained for a number of epochs (x-axis) on the ImageNet dataset. Vertical bars indicate the step decays of the learning rate. Triangles indicate the epochs corresponding to the highest success rate per target.}
\label{fig:plot_successrate_epochs_standard_schedule_imagenet}
\end{figure}

\begin{figure}[h]
\begin{center}
   \includegraphics[width=0.7\linewidth]{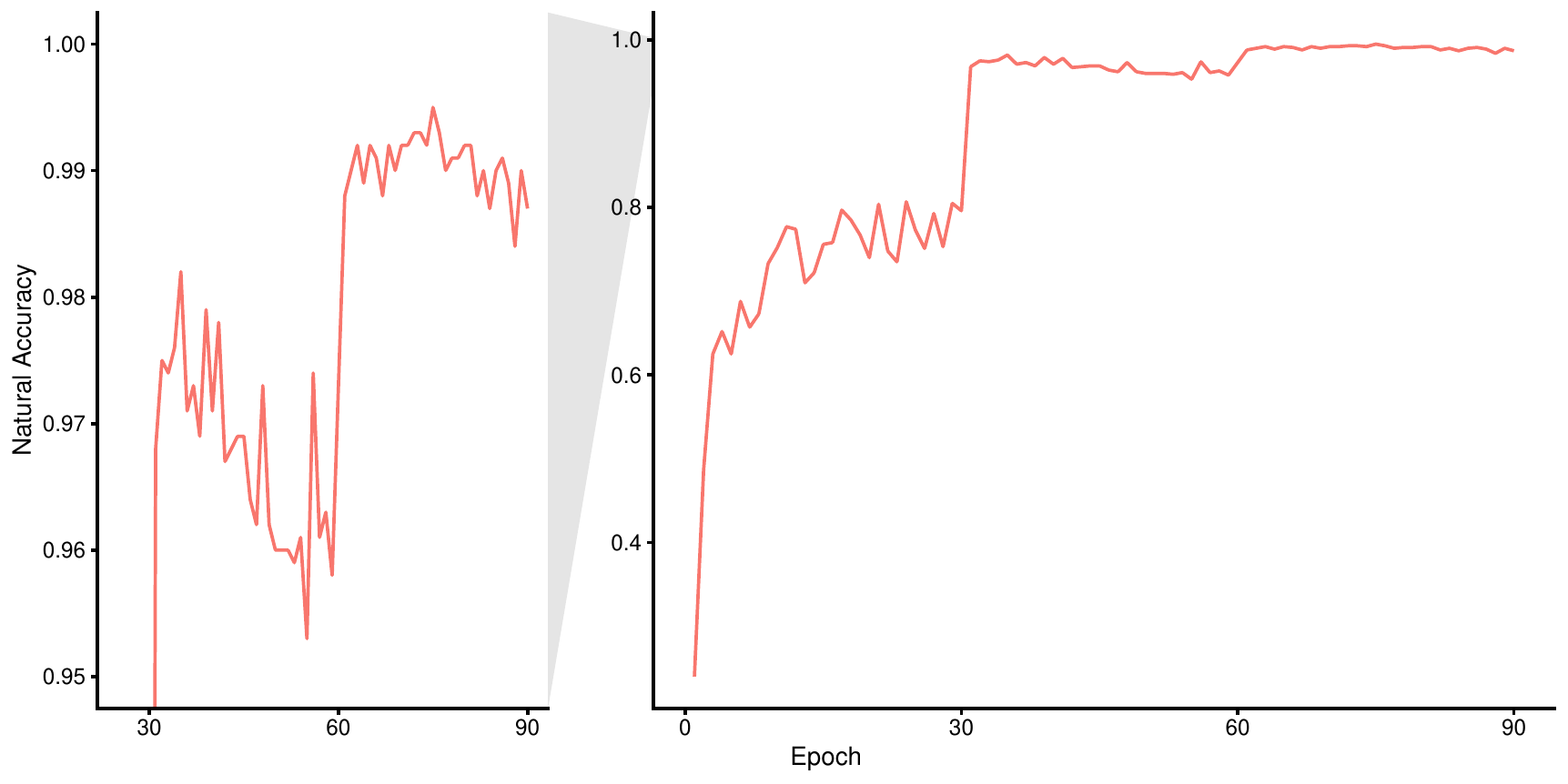}
\end{center}
   \caption{Natural test accuracy of the ResNet-50 surrogate model trained for a number of epochs (x-axis) on the ImageNet dataset. Evaluated on the test subset used to craft adversarial examples in Figure \ref{fig:plot_successrate_epochs_standard_schedule_imagenet}.}
\label{fig:plot_acc_epochs_standard_imagenet}
\end{figure}

Early stopping clearly benefits transferability for all ten targets on CIFAR-10 and all ten targets on ImageNet (except for the two Vision Transformers, where the transferability plateaus). We reproduce below the success rates for all target models from the ResNet-50 surrogate model on both CIFAR-10 (Figure \ref{fig:plot_successrate_epochs_standard_schedule}) and ImageNet (Figure \ref{fig:plot_successrate_epochs_standard_schedule_imagenet}) datasets. We also report the evolution of the natural accuracy for both CIFAR-10 (Figure \ref{fig:plot_acc_epochs_standard}) and ImageNet (Figure \ref{fig:plot_acc_epochs_standard_imagenet}).

\clearpage
\section{Another Look at the Non-Robust Features Hypothesis about Early Stopping}
\label{sec:app-rfs-nrfs}

\begin{figure}[h]
\begin{center}
   \includegraphics[width=\linewidth]{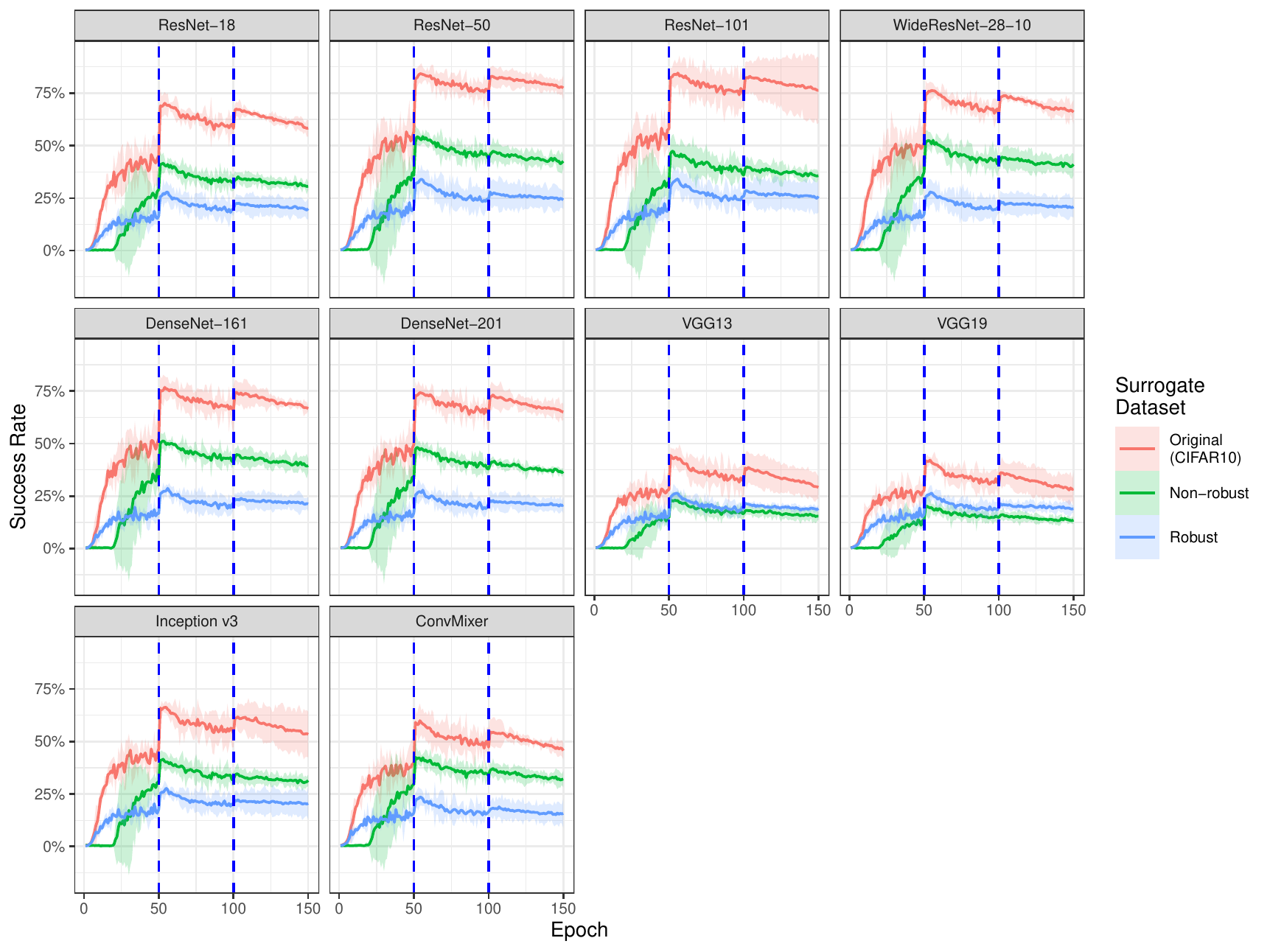}
\end{center}
   \caption{Early stopping improves transferability of surrogate models trained on both robust and non-robust datasets. Success rate evaluated over ten target models (title subfigure) from a ResNet-50 surrogate model trained for a number of epochs (x-axis) on the datasets $D_R$ (blue) and $D_\text{NR}$ (green) of \cite{Ilyas2019AdversarialFeatures} modified from CIFAR-10 (red).}
\label{fig:xp_rfs_nrfs_plot_dr_dnr}
\end{figure}

This section contains detailed results of Section 3. Figure \ref{fig:xp_rfs_nrfs_plot_dr_dnr} reports the transferability per target of the experiment that shows the success of early stopping for surrogates trained on both robust and non-robust datasets. For this experiment, we divided by two the initial learning rate ($0.05$) when training on $D_\text{NR}$ due to instabilities during training when trained with a learning rate of $0.1$.

\clearpage
\section{Transferability and Training Dynamics}
\label{sec:app-training-dynamics}

This section contains additional results of Section 4 on the relationship between the training dynamics of the surrogate model and its transferability. 

\subsection{Consistency of the Peak of Transferability}

\begin{figure}[h]
    \begin{center}
        \includegraphics[width=\linewidth]{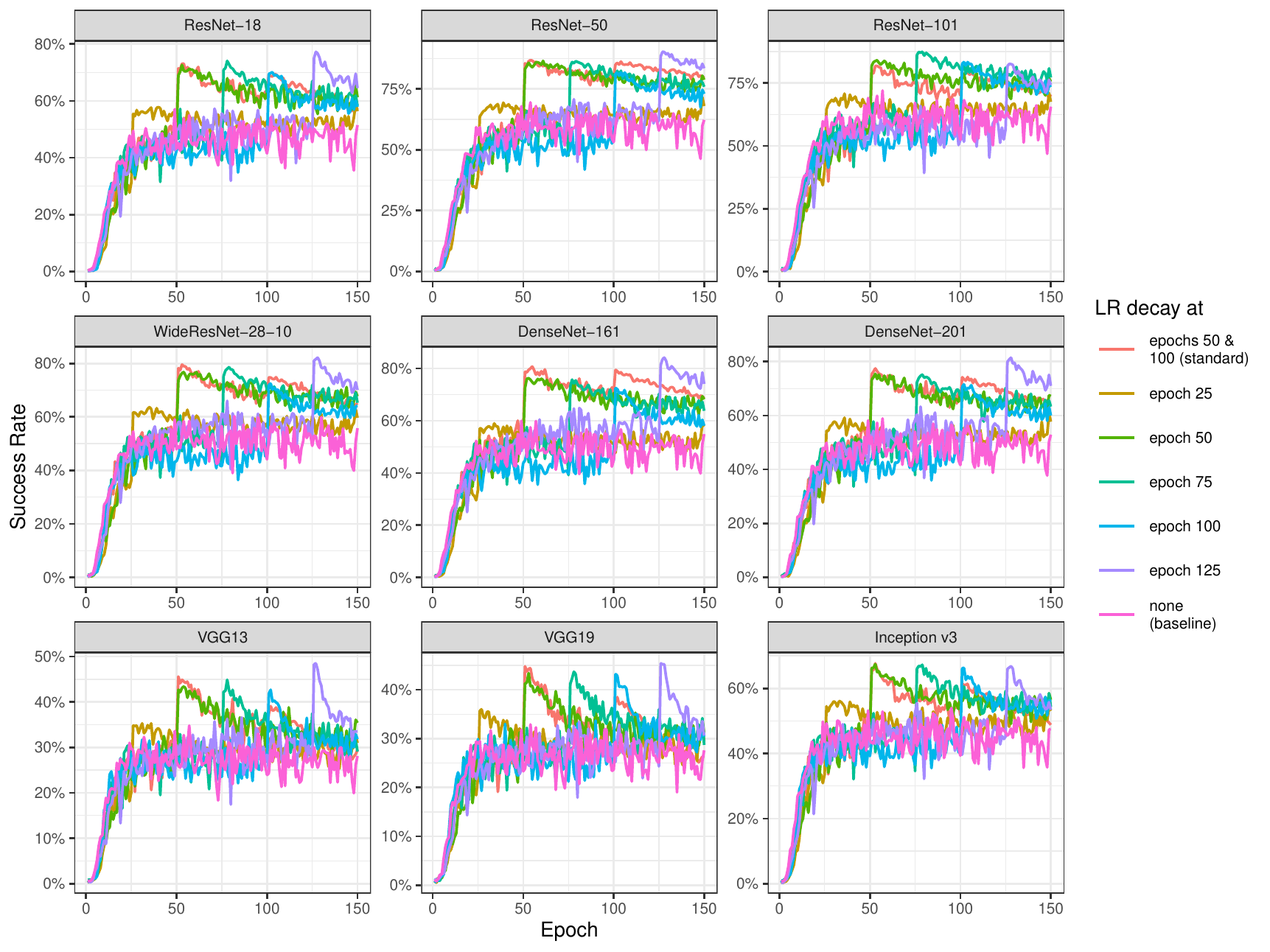}
    \end{center}
   \caption{Step learning rate decay benefits transferability at any epochs. Success rate evaluated over nine target models (title subfigure) from a ResNet-50 surrogate model trained for a number of epochs (x-axis) on the CIFAR-10. The LR is divided by 10 a single time during training at an epoch indicated by the color. Scale not shared between subfigures.}
    \label{fig:xp_lr_single_decay_plot_successrate_epochs_single_decay}
\end{figure}

Figure \ref{fig:xp_lr_single_decay_plot_successrate_epochs_single_decay} contains the transferability per target of the surrogate models trained with a single learning rate decay at a varying epoch. The consistency of the peak of transferability across training epochs is valid for all ten targets.

\begin{figure}[h]
    \begin{center}
        \includegraphics[width=0.4\linewidth]{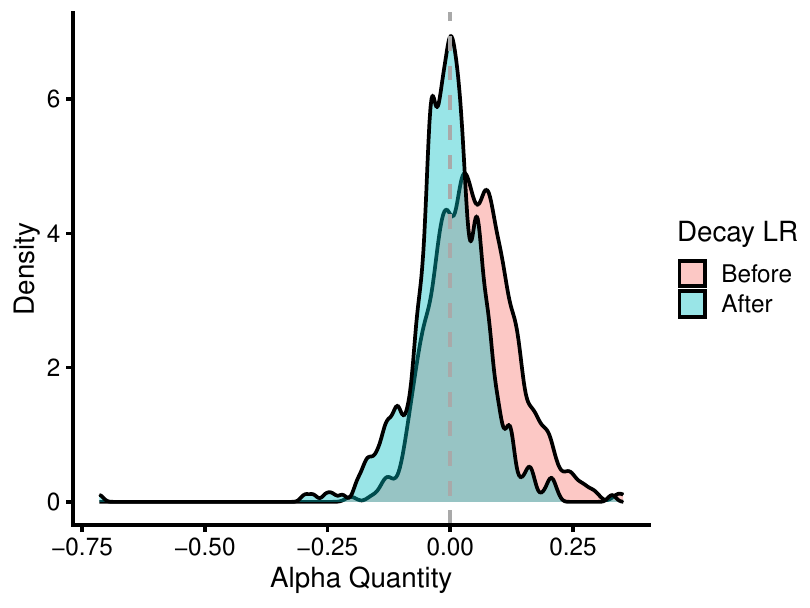}
    \end{center}
   \caption{The LR decay corresponds to a transition from a “crossing the
valley” phase to a “crawling down to the valley” phase. Density plot of the $\alpha$-quantity values computed each four SGD iterations during the best five epochs for transferability on CIFAR-10 (epochs 50--54, ``After'' group, blue) and the five preceding epochs (epochs 45--49, ``Before'' group, red).}
    \label{fig:plot_alpha_density}
\end{figure}

\subsection{Crossing the Valley Before Exploring the Valley}

Before the learning rate decays, the exploration tends to behave more like ``crossing the valley'' than after decay, when it is more likely to ``crawl down to the valley'', as described in \cite{Schneider2021Cockpit:Networks}. Figure \ref{fig:illustration} illustrates this phenomenon. \cite{Schneider2021Cockpit:Networks} proposes the $\alpha$-quantity, a metric computed at the level of SGD iterations to disentangle whether the iteration understeps or overshoots the minimum along the current step direction. Based on a noise-informed quadratic fit,  $\alpha \approx 0$ indicates an appropriate LR that minimizes the loss in the direction of the gradient at this iteration (``going down to the valley''). $\alpha > 0$ indicates that the current LR overshoots this minimum (``crossing the valley''). We compute the $\alpha$-quantity every four SGD iterations during the best five epochs for transferability on CIFAR-10 (``after LR decay'', epochs 50--54) and during the five preceding epochs (``before LR decay'', epochs 45--49). The one-sided Welch Two Sample t-test has a p-value inferior to $2.2\mathrm{e}^{-16}$. We reject the null hypothesis in favor of the alternative hypothesis that the true difference of $\alpha$-quantity in means between the group ``before LR decay'' and the group ``after LR decay'' is strictly greater than 0. We also perform a one-sided Welch Two Sample t-test on the 5 epochs before and after the second LR decay (epochs 95--99 vs. epochs 100-105). Its p-value is equal to $0.004387$. Using the Bonferroni correction, we compare the p-values of both individual tests with a significance threshold of 0.5\%. We reject the null hypothesis for both LR decays with a significance level of 1\%. Figure \ref{fig:plot_alpha_density} is the density plot of the $\alpha$-quantities for both groups. Our results suggest that before the LR decay, training is slow due to a ``crossing the valley'' pattern. The best early stopped surrogate occurs a few training epochs after the LR decay when the SGD starts exploring the bottom of the valley.

\clearpage
\section{Transferability from SAM and Its Variants}
\label{sec:app-sam-hp}

This section presents the following elements:
\begin{enumerate}
    \item A description of SAM and its variants (Section \ref{sec:app-sam-description}),
    \item The success rate with respect to the $\rho$ hyperparameter of SAM and its variants, used to tune this hyperparameter for transferability (Section \ref{sec:app-sam-hp-subsection}),
    \item The natural accuracy of the surrogates trained by SAM and its variants (Section \ref{sec:app-sam-acc}).
\end{enumerate}

\subsection{Description of SAM and its variants}
\label{sec:app-sam-description}

\begin{figure}[h]
    \begin{center}
        \includegraphics[width=0.45\linewidth]{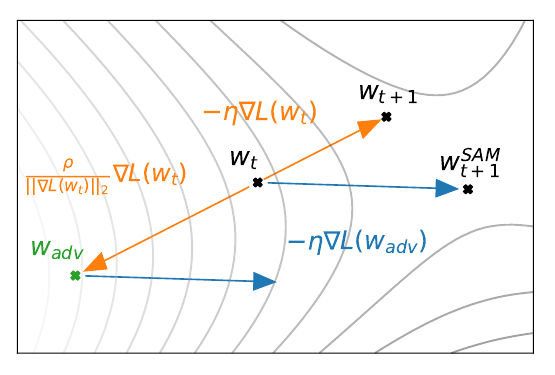}
    \end{center}
   \caption{Illustrative schema of a training iteration with SAM. Illustration from \citet{Foret2020Sharpness-AwareGeneralization}.}
    \label{fig:schema-sam}
\end{figure}

This section describes SAM and its variants. It includes an illustrative schema of SAM (Figure \ref{fig:schema-sam}).

\begin{description}
    \item[SAM] SAM \citep{Foret2020Sharpness-AwareGeneralization} minimizes the maximum loss around a neighborhood by performing a gradient ascent step. First, the gradient ascent step is performed (left orange in Figure \ref{fig:schema-sam}) to compute $\epsilon_t = \rho \frac{\nabla \mathcal{L}(w_t)}{\|\nabla \mathcal{L}(w_t)\|_2}$. It is followed by a gradient descent step, $w_{t+1} = w_t - \alpha_t \left( \nabla \mathcal{L}(w_t + \epsilon_t) + \lambda w_t \right)$ (blue arrows in Figure \ref{fig:schema-sam}). 
    \item[LookSAM] LookSAM \citep{liuEfficientScalableSharpnessAware2022} is an efficient alternative that computes only the additional ascending gradient of SAM once per five training iterations, i.e., $\epsilon_{t} = \epsilon_{\left\lfloor \frac{t}{5} \right\rfloor}$. We faced some convergence issues when applying it with our learning rate schedule (the original authors used a schedule with warmup). To solve this issue, we add some warmup: LookSAM computes both gradients for the first three epochs of training, exactly as SAM. From the fourth epoch, the training resumes to the efficient LookSAM variant. The computational cost reported in Table \ref{tab:competitive-ttechs-new} takes into account this overhead.
    \item[ASAM] ASAM \citep{Kwon2021ASAM:Networks} is an adaptive variant of SAM. ASAM introduces a normalization operator $T^{-1}_{w}$ to adjust the maximization region with respect to the weight scale. The maximization step becomes: $\epsilon_t = \rho \frac{T^2_{w_t} \nabla \mathcal{L}_s(w_t)}{\|T_{w_t} \nabla \mathcal{L}_s(w_t)\|_2}$. We follow the original paper \cite{Kwon2021ASAM:Networks} to select the hyperparameter $\rho$: the authors recommend multiplying $\rho$ by 10 when switching to an adaptive variant.
    \item[GSAM] GSAM \citep{Zhuang2022SurrogateTraining} introduces a decomposition of the gradient computed in the maximization step. Only the orthogonal component is then used in the minimization step. AGSAM is the combination of GSAM and ASAM.
\end{description}

\subsection{The size of flat neighborhoods: the choice of the $\rho$ hyperparameter}
\label{sec:app-sam-hp-subsection}

\begin{figure}[h]
    \begin{center}
        \includegraphics[width=\linewidth]{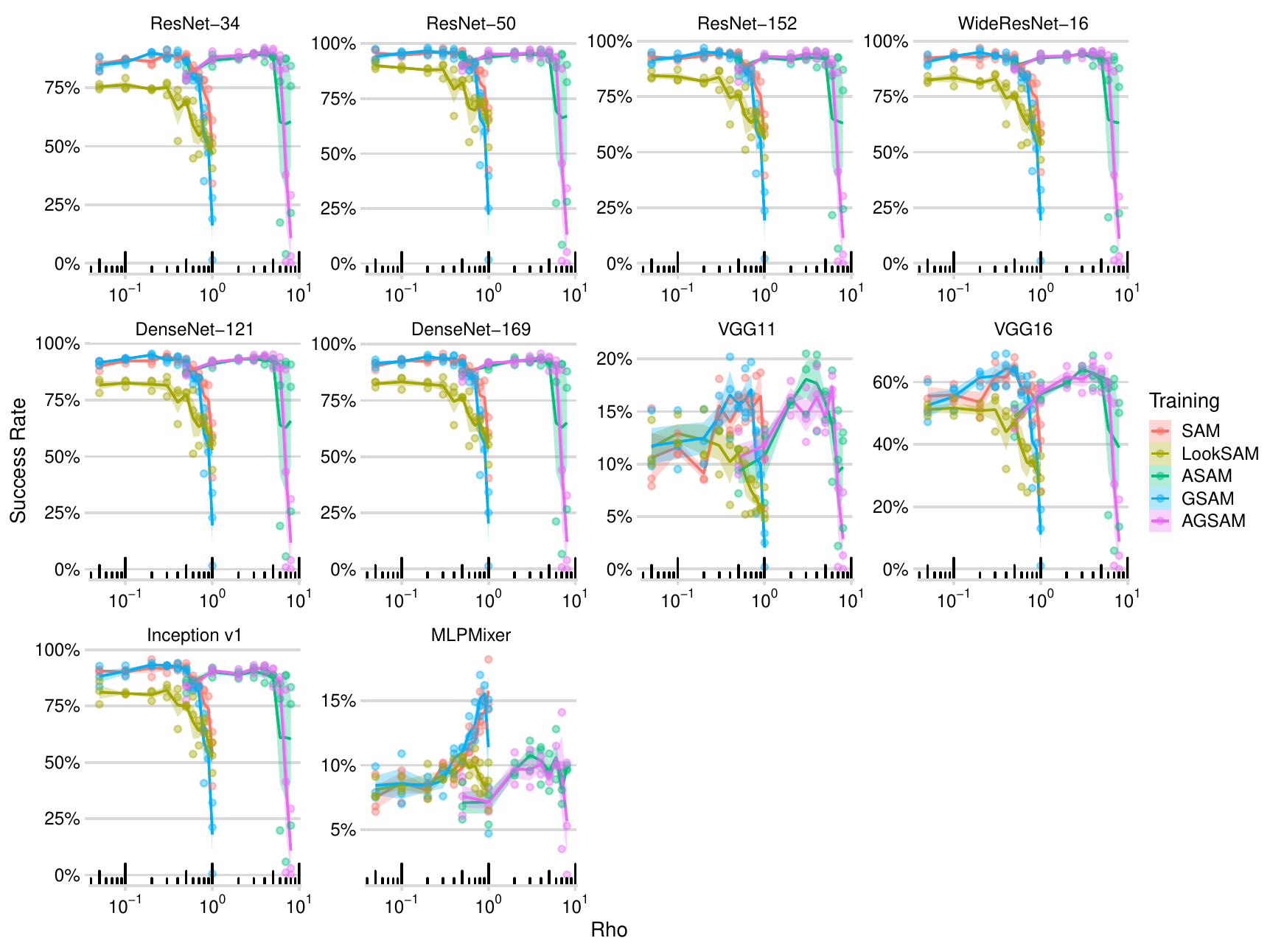}
    \end{center}
   \caption{All SAM variants trains better surrogate models with a larger $\rho$ than the original one, used for natural accuracy. \textit{Validation} success rate on ten validation target models (subfigure title) from a ResNet-50 surrogate model trained using SAM or SAM variants (colors) with various $\rho$ hyperparameters (x-axis) on the CIFAR-10 dataset. Adversarial examples are crafted from a disjoint subset of one thousand original examples from the train set. The left most $\rho$ value is the original one: 0.5 for adaptive variants (ASAM, AGSAM), 0.05 for others. Average (line) and $\pm$ one standard deviation (colored area) of three training runs.}
    \label{fig:plot_hp_rho_successrate_epochs_val_3seeds}
\end{figure}

A stronger regularization induced by SAM with large flat neighborhoods trains a better surrogate model. The size of flat neighborhoods is controlled by the unique hyperparameter of SAM, noted $\rho$. Figure \ref{fig:plot_hp_rho_successrate_epochs_val_3seeds} reports the validation success rate used to find the best large $\rho$ for each SAM variants. The selected $\rho$ values are reported for each SAM variant in Table \ref{tab:hp-training}. This success rate is computed on a separate set of target models, surrogate models, and a set of examples. This experimental setting is carefully designed to avoid data leakage by optimizing the hyperparameter against specific target models. Otherwise, this could result in model selection, similar to query-based attacks, which are not allowed by our threat model of transfer-based black-box attacks.

\begin{figure}[h]
    \begin{center}
        \includegraphics[width=\linewidth]{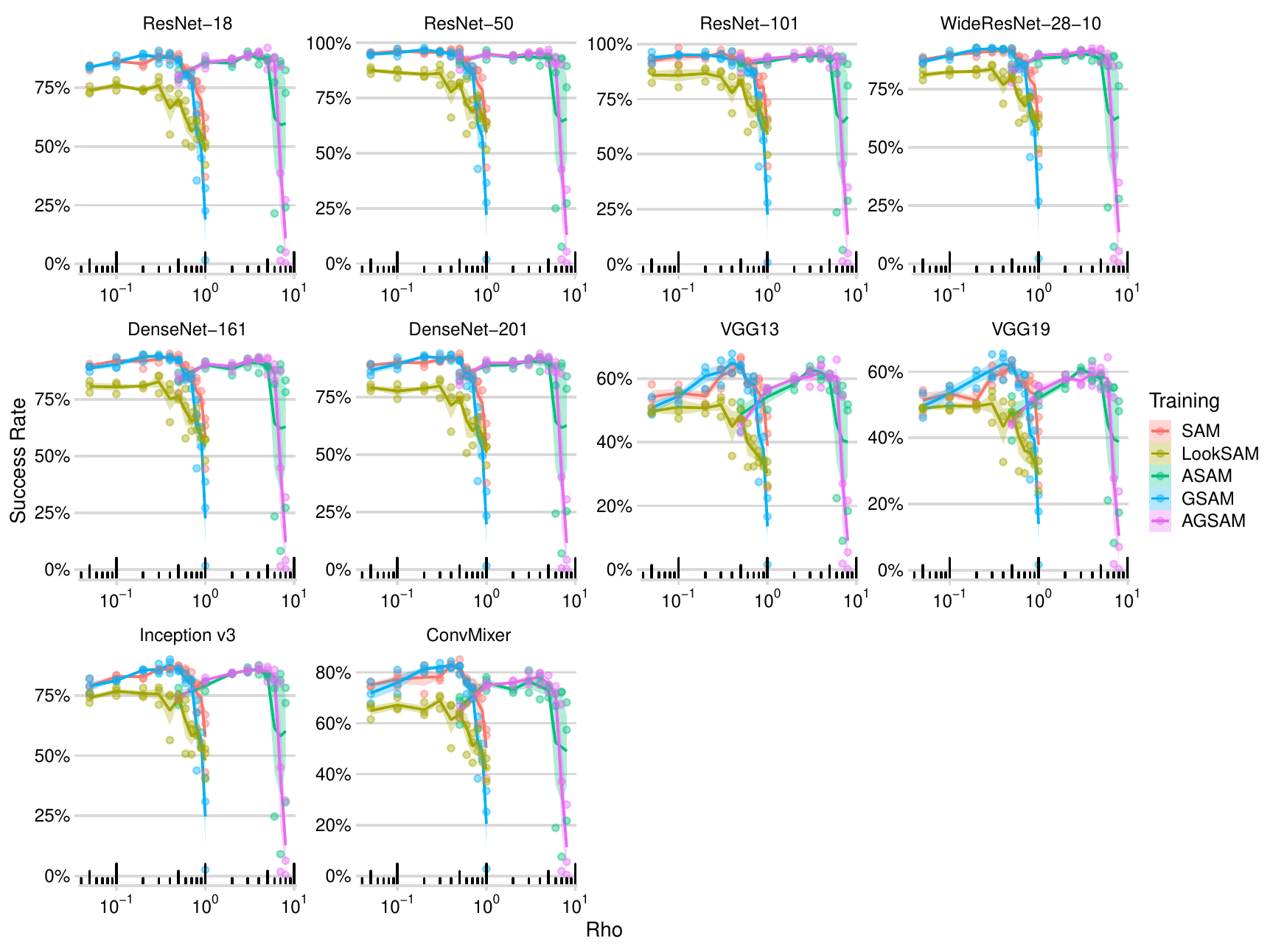}
    \end{center}
   \caption{All SAM variants trains better surrogate models with a larger $\rho$ than the original one, used for natural accuracy. \textit{Test} success rate on ten test target models (subfigure title) from a ResNet-50 surrogate model trained using SAM or SAM variants (colors) with various $\rho$ hyperparameters (x-axis) on the CIFAR-10 dataset. Adversarial examples are crafted from a disjoint subset of one thousand original examples from the test set. The left most $\rho$ value is the original one: 0.5 for adaptive variants (ASAM, AGSAM), 0.05 for others. Average (line) and $\pm$ one standard deviation (colored area) of three training runs. The surrogate models used here are the same as in Figure \ref{fig:plot_hp_rho_successrate_epochs_val_3seeds}. Nevertheless, Sections \ref{sec:sam} and \ref{sec:evalution} report results from three other independently trained surrogate models.}
    \label{fig:plot_hp_rho_successrate_epochs_test_3seeds}
\end{figure}

Figure \ref{fig:plot_hp_rho_successrate_epochs_test_3seeds} reports the test success rate on the same surrogate models, but computed on our test set of target models and using natural examples from the test set. Sections \ref{sec:sam} and \ref{sec:evalution} report results from three other independently trained surrogate models. The transferability improvement of LookSAM with large $\rho$ is tiny compared to LookSAM with the original $\rho$. LookSAM is an efficient variant of SAM that skips 4/5 of the additional ascending gradients of SAM. Our hypothesis is that training with large $\rho$ requires a more refined update strategy.

\clearpage

\subsection{Natural Accuracy of SAM and Its Variants}
\label{sec:app-sam-acc}

Tables \ref{tab:accuracy_sam_imagenet} and \ref{tab:accuracy_sam_cifar10} report the natural test accuracies of the surrogate models studied in Sections \ref{sec:sam} and \ref{sec:evalution}. As commented in Section \ref{sec:sam}, the strong regularization induced by SAM with large flat neighborhoods (high $\rho$) can degrade natural generalization. In particular, on ImageNet, our ResNet-18 and ResNet-50 surrogates trained with l-SAM have a worst natural accuracy compared to SAM and even fully trained SGD. On CIFAR-10, l-SAM has an inferior natural accuracy than SAM, and a similar one to SGD. Therefore, the improvement in transferability from l-SAM, i.e., \textit{the generalization of adversarial examples from this strong regularization, cannot be explained by an improvement in natural generalization}, i.e, a better fit to the data.

\begin{table}[h]
\begin{center}
\caption{Accuracy computed on the test set of the surrogates trained by SAM and its variants on ImageNet. In \%.}
\label{tab:accuracy_sam_imagenet}

\begin{tabular}[t]{lllr}
\toprule
Arch & Training & Size
neighborhood & Accuracy\\
\midrule
\cellcolor{gray!6}{ResNet-18} & \cellcolor{gray!6}{SGD (baseline)} & \cellcolor{gray!6}{None}
\cellcolor{gray!6}{(SGD)} & \cellcolor{gray!6}{69.8}\\
ResNet-18 & SAM & Large & 67.9\\
\cellcolor{gray!6}{ResNet-18} & \cellcolor{gray!6}{SAM} & \cellcolor{gray!6}{Original} & \cellcolor{gray!6}{70.3}\\
ResNet-18 & GSAM & Large & 68.8\\
\cellcolor{gray!6}{ResNet-18} & \cellcolor{gray!6}{GSAM} & \cellcolor{gray!6}{Original} & \cellcolor{gray!6}{70.3}\\
ResNet-18 & ASAM & Large & 68.9\\
\cellcolor{gray!6}{ResNet-18} & \cellcolor{gray!6}{ASAM} & \cellcolor{gray!6}{Original} & \cellcolor{gray!6}{70.2}\\
ResNet-18 & AGSAM & Large & 67.8\\
\cellcolor{gray!6}{ResNet-18} & \cellcolor{gray!6}{AGSAM} & \cellcolor{gray!6}{Original} & \cellcolor{gray!6}{70.1}\\
ResNet-50 & SGD
(baseline) & None
(SGD) & 75.7\\
\cellcolor{gray!6}{ResNet-50} & \cellcolor{gray!6}{SAM} & \cellcolor{gray!6}{Large} & \cellcolor{gray!6}{74.5}\\
\bottomrule
\end{tabular}
\end{center}
\end{table}

\begin{table}[h]
\begin{center}
\caption{Accuracy computed on the test set of the surrogates trained by SAM and its variants on CIFAR-10. In \%.}
\label{tab:accuracy_sam_cifar10}

\begin{tabular}[t]{lllr}
\toprule
Arch & Training & Size neighborhood & Accuracy\\
\midrule
\cellcolor{gray!6}{ResNet-50} & \cellcolor{gray!6}{SGD (baseline)} & \cellcolor{gray!6}{None (SGD)} & \cellcolor{gray!6}{94.5 ±0.4}\\
ResNet-50 & SAM & Large & 94.6 ±0.4\\
\cellcolor{gray!6}{ResNet-50} & \cellcolor{gray!6}{SAM} & \cellcolor{gray!6}{Original} & \cellcolor{gray!6}{95.3 ±0.3}\\
ResNet-50 & GSAM & Large & 94.7 ±0.5\\
\cellcolor{gray!6}{ResNet-50} & \cellcolor{gray!6}{GSAM} & \cellcolor{gray!6}{Original} & \cellcolor{gray!6}{95.4 ±0.5}\\
ResNet-50 & ASAM & Large & 95.6 ±0.5\\
\cellcolor{gray!6}{ResNet-50} & \cellcolor{gray!6}{ASAM} & \cellcolor{gray!6}{Original} & \cellcolor{gray!6}{95.1 ±0.4}\\
ResNet-50 & AGSAM & Large & 95.9 ±0.3\\
\cellcolor{gray!6}{ResNet-50} & \cellcolor{gray!6}{AGSAM} & \cellcolor{gray!6}{Original} & \cellcolor{gray!6}{95.3 ±0.6}\\
\bottomrule
\end{tabular}
\end{center}
\end{table}


\clearpage
\section{Transferability and Weight Decay}
\label{sec:app-wd}

We show that in the case of weight decay, a stronger regularization of the surrogate model does not improve transferability. Unlike weight decay, the stronger regularization of SAM is tightly linked to transferability. 

\begin{figure}[h]
    \begin{center}
    \includegraphics[width=0.7\linewidth]{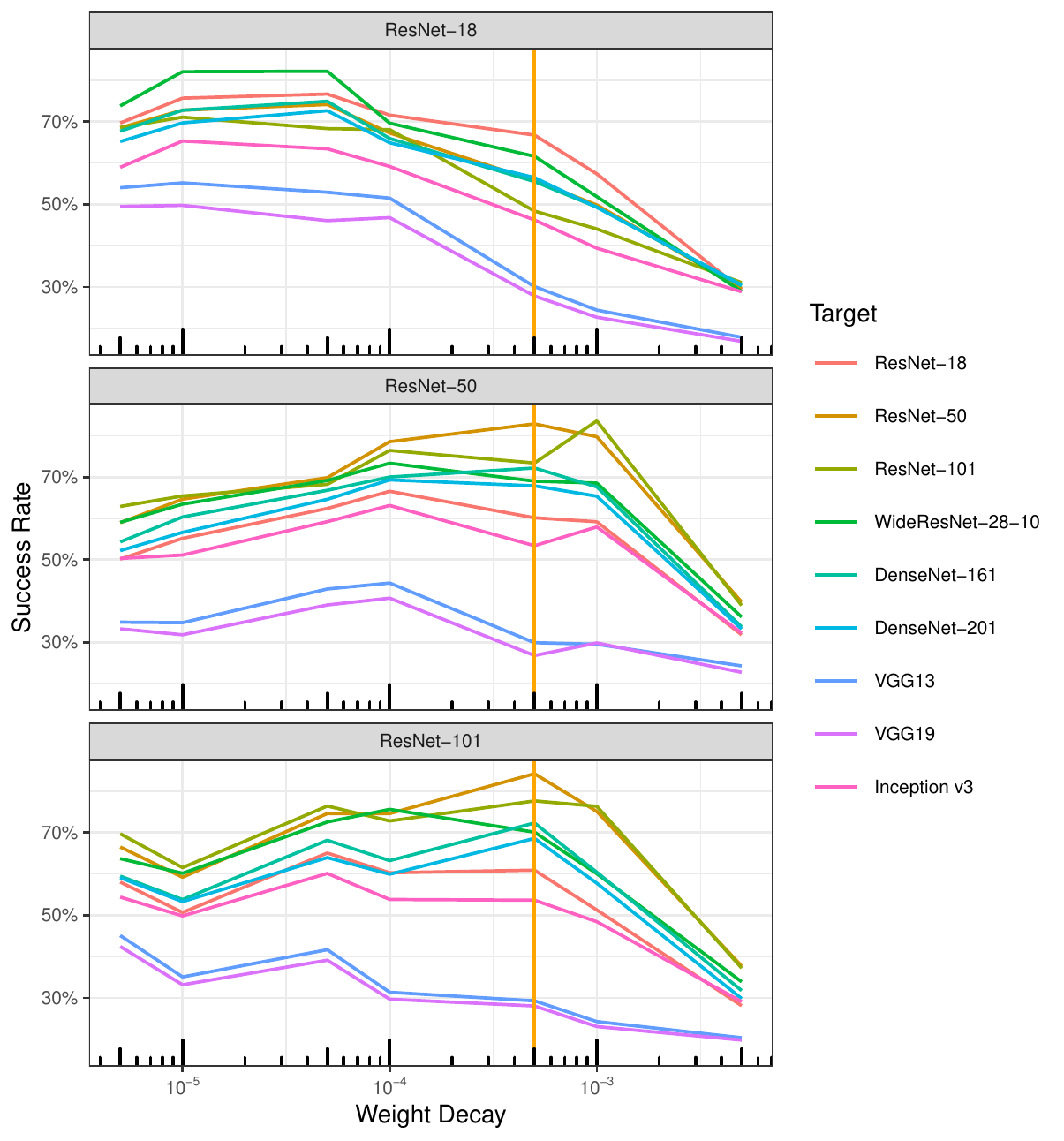}
    \end{center}
    \caption{Stronger weight decay regularization does not improve transferability. Success rate from ResNet surrogates (subfigure title) trained with a weight decay (x-axis) evaluated on targets (colors) trained with weight decay indicated by the yellow vertical bar on the CIFAR-10 dataset.}
    \label{fig:transferability-wd}
\end{figure}

We train on CIFAR-10 one surrogate model for various values of the weight decay regularization (5e-3, 1e-3, 5e-4, 1e-4, 5e-5, 1e-5 and 5e-6) and for various capacities of the ResNet architecture (ResNet-18, ResNet-50, ResNet-101). Figure \ref{fig:transferability-wd} presents the transferability of these surrogates. For the ResNet-50 and ResNet-101 surrogates, the best average success rate simply corresponds to the weight decay used to train the target models. Interestingly, a lighter weight decay regularization trains better ResNet-18 surrogate models. We hypothesize that a ligher regularization allows this smaller architecture to better mimic the complexities of the larger architectures used as targets. Overall, a stronger weight decay regularization does not train better surrogate models, contrary to the SAM regularization.

\clearpage
\section{Evaluation of l-SAM: Improving Transferability Techniques With Sharpness Minimization}
\label{sec:app-eval-rfn}

This section extends the evaluation of SAM with large flat neighborhoods of Section \ref{sec:evalution}, performed with $\varepsilon$ equal to $4/255$, to two other perturbation $L_\infty$ norms ($2/255$ and $8/255$) for the competitive techniques, and reports the success rate per target for the complementary techniques.

\paragraph{Evaluation against competitive techniques.}  Tables \ref{tab:competitive-ttechs-cifar-eps2} and \ref{tab:competitive-ttechs-cifar-eps8} evaluate competitive techniques of l-SAM on CIFAR-10 with, respectively, maximum perturbations $L_\infty$ norm $\varepsilon$ of $2/255$ and $8/255$. The same conclusions made with perturbations of size $4/255$ hold for these two norms: l-SAM clearly improves transferability. l-SAM beats other competitive techniques for the ten targets and both norms. Tables \ref{tab:competitive-ttechs-imagenet-eps2}, \ref{tab:competitive-ttechs-imagenet}, and \ref{tab:competitive-ttechs-imagenet-eps8} show, respectively, that l-SAM beats the other techniques in 6 out of 10 targets for $\varepsilon$ equal to $2/255$, and in 5 out of 10 targets for $\varepsilon$ equal to $8/255$.

\paragraph{Evaluation with complementary techniques.} Tables \ref{tab:table_main_results_complementary_ttechimagenet_eps2}, \ref{tab:table_main_results_complementary_ttechimagenet}, and  \ref{tab:table_main_results_complementary_ttechimagenet_eps8} report in detail per target the evaluation of complementary transferability techniques on ImageNet. l-SAM increases the transferability of every nine techniques against every ten targets when combined, except for LGV on 4 targets using $\varepsilon$ equals $2/255$, and LGV on 3 targets with $\varepsilon$ equals $4/255$ or $8/255$. Since LGV collects models with SGD and a high learning rate, a conflict might occur when LGV continues training with SGD from a checkpoint trained with SAM. Future work may explore the adaptation of the LGV model collection to SAM.

\begin{table*}[h]
\begin{center}
\caption{Success rate on CIFAR-10 of competitive techniques to train a single surrogate model. Adversarial examples evaluated on nine targets with a maximum perturbation $L_\infty$ norm $\varepsilon$ of $2/255$. Bold is best. In \%.}

\resizebox{\textwidth}{!}{%
\begin{tabular}[t]{lrrrrrrrrr}
\toprule
\multicolumn{1}{c}{ } & \multicolumn{9}{c}{Target} \\
\cmidrule(l{3pt}r{3pt}){2-10}
Surrogate & RN18 & RN50 & RN101 & DN161 & DN201 & VGG13 & VGG19 & IncV3 & WRN28\\
\midrule
\cellcolor{gray!6}{Fully Trained SGD} & \cellcolor{gray!6}{24.2} & \cellcolor{gray!6}{44.7} & \cellcolor{gray!6}{35.6} & \cellcolor{gray!6}{33.3} & \cellcolor{gray!6}{31.4} & \cellcolor{gray!6}{9.6} & \cellcolor{gray!6}{9.2} & \cellcolor{gray!6}{22.6} & \cellcolor{gray!6}{30.8}\\
Early Stopped SGD & 28.6 & 46.1 & 38.6 & 36.3 & 34.6 & 12.7 & 13.0 & 27.1 & 34.9\\
\cellcolor{gray!6}{SAT} & \cellcolor{gray!6}{19.7} & \cellcolor{gray!6}{27.3} & \cellcolor{gray!6}{25.4} & \cellcolor{gray!6}{20.1} & \cellcolor{gray!6}{20.3} & \cellcolor{gray!6}{13.4} & \cellcolor{gray!6}{13.5} & \cellcolor{gray!6}{17.6} & \cellcolor{gray!6}{20.5}\\
l-SAM (ours) & \textbf{45.4} & \textbf{67.1} & \textbf{60.6} & \textbf{58.9} & \textbf{55.8} & \textbf{20.5} & \textbf{19.8} & \textbf{45.0} & \textbf{54.1}\\
\bottomrule
\end{tabular}
}

\label{tab:competitive-ttechs-cifar-eps2}
\end{center}
\end{table*}

\begin{table*}
\begin{center}
\caption{Success rate on CIFAR-10 of competitive techniques to train a single surrogate model. Adversarial examples evaluated on nine targets with a maximum perturbation $L_\infty$ norm $\varepsilon$ of $4/255$. Bold is best. In \%.}

\resizebox{\textwidth}{!}{%
\begin{tabular}[t]{lrrrrrrrrr}
\toprule
\multicolumn{1}{c}{ } & \multicolumn{9}{c}{Target} \\
\cmidrule(l{3pt}r{3pt}){2-10}
Surrogate & RN18 & RN50 & RN101 & DN161 & DN201 & VGG13 & VGG19 & IncV3 & WRN28\\
\midrule
\cellcolor{gray!6}{Fully Trained SGD} & \cellcolor{gray!6}{57.9} & \cellcolor{gray!6}{81.2} & \cellcolor{gray!6}{70.6} & \cellcolor{gray!6}{70.8} & \cellcolor{gray!6}{66.1} & \cellcolor{gray!6}{27.8} & \cellcolor{gray!6}{26.3} & \cellcolor{gray!6}{49.4} & \cellcolor{gray!6}{66.5}\\
Early Stopped SGD & 73.3 & 87.8 & 82.1 & 81.4 & 78.3 & 45.5 & 44.3 & 66.8 & 79.5\\
\cellcolor{gray!6}{SAT} & \cellcolor{gray!6}{66.3} & \cellcolor{gray!6}{76.2} & \cellcolor{gray!6}{73.6} & \cellcolor{gray!6}{66.9} & \cellcolor{gray!6}{66.1} & \cellcolor{gray!6}{49.8} & \cellcolor{gray!6}{48.5} & \cellcolor{gray!6}{57.9} & \cellcolor{gray!6}{67.8}\\
l-SAM \textbf{(ours)} & \textbf{89.7} & \textbf{97.3} & \textbf{95.5} & \textbf{95.7} & \textbf{94.0} & \textbf{63.6} & \textbf{60.6} & \textbf{87.3} & \textbf{93.0}\\

\bottomrule
\end{tabular}
}

\label{tab:competitive-ttechs-cifar}
\end{center}
\end{table*}

\begin{table*}[h]
\begin{center}
\caption{Success rate on CIFAR-10 of competitive techniques to train a single surrogate model. Adversarial examples evaluated on nine targets with a maximum perturbation $L_\infty$ norm $\varepsilon$ of $8/255$. Bold is best. In \%.}
\resizebox{\textwidth}{!}{%
\begin{tabular}[t]{lrrrrrrrrr}
\toprule
\multicolumn{1}{c}{ } & \multicolumn{9}{c}{Target} \\
\cmidrule(l{3pt}r{3pt}){2-10}
Surrogate & RN18 & RN50 & RN101 & DN161 & DN201 & VGG13 & VGG19 & IncV3 & WRN28\\
\midrule
\cellcolor{gray!6}{Fully Trained SGD} & \cellcolor{gray!6}{88.3} & \cellcolor{gray!6}{97.4} & \cellcolor{gray!6}{92.4} & \cellcolor{gray!6}{93.9} & \cellcolor{gray!6}{91.4} & \cellcolor{gray!6}{64.2} & \cellcolor{gray!6}{60.5} & \cellcolor{gray!6}{79.3} & \cellcolor{gray!6}{91.9}\\
Early Stopped SGD & 97.8 & 99.6 & 98.8 & 98.9 & 98.4 & 89.1 & 87.5 & 95.6 & 98.8\\
\cellcolor{gray!6}{SAT} & \cellcolor{gray!6}{97.0} & \cellcolor{gray!6}{98.7} & \cellcolor{gray!6}{98.0} & \cellcolor{gray!6}{97.1} & \cellcolor{gray!6}{96.4} & \cellcolor{gray!6}{90.2} & \cellcolor{gray!6}{89.2} & \cellcolor{gray!6}{93.2} & \cellcolor{gray!6}{97.1}\\
l-SAM (ours) & \textbf{99.7} & \textbf{100.0} & \textbf{100.0} & \textbf{100.0} & \textbf{99.9} & \textbf{96.6} & \textbf{95.6} & \textbf{99.6} & \textbf{99.9}\\
\bottomrule
\end{tabular}
}

\label{tab:competitive-ttechs-cifar-eps8}
\end{center}
\end{table*}

\begin{table*}[h]
\begin{center}
\caption{Success rate on ImageNet of competitive techniques to train a single surrogate model. Adversarial examples evaluated on ten targets with a maximum perturbation $L_\infty$ norm $\varepsilon$ of $2/255$. Bold is best. In \%.}

\resizebox{\textwidth}{!}{%
\begin{tabular}[t]{lrrrrrrrrrr}
\toprule
\multicolumn{1}{c}{ } & \multicolumn{10}{c}{Target} \\
\cmidrule(l{3pt}r{3pt}){2-11}
Surrogate & RN50 & RN152 & RNX50 & WRN50 & VGG19 & DN201 & IncV1 & IncV3 & ViT B & SwinS\\
\midrule
\cellcolor{gray!6}{Fully Trained SGD} & \cellcolor{gray!6}{18.7} & \cellcolor{gray!6}{9.4} & \cellcolor{gray!6}{10.0} & \cellcolor{gray!6}{9.3} & \cellcolor{gray!6}{7.6} & \cellcolor{gray!6}{5.8} & \cellcolor{gray!6}{4.8} & \cellcolor{gray!6}{5.2} & \cellcolor{gray!6}{1.1} & \cellcolor{gray!6}{1.4}\\
Early Stopped SGD & 23.8 & 10.7 & 10.6 & 10.6 & 8.7 & 6.8 & 5.6 & 6.1 & 1.1 & 1.5\\
\cellcolor{gray!6}{LGV-SWA} & \cellcolor{gray!6}{49.3} & \cellcolor{gray!6}{24.8} & \cellcolor{gray!6}{25.0} & \cellcolor{gray!6}{21.7} & \cellcolor{gray!6}{18.5} & \cellcolor{gray!6}{16.8} & \cellcolor{gray!6}{11.6} & \cellcolor{gray!6}{7.9} & \cellcolor{gray!6}{1.4} & \cellcolor{gray!6}{1.5}\\
SAT & 30.0 & 19.2 & 24.4 & 20.6 & 18.4 & 20.2 & \textbf{20.0} & \textbf{16.6} & \textbf{4.9} & \textbf{4.4}\\
\cellcolor{gray!6}{l-SAM (ours)} & \cellcolor{gray!6}{\textbf{53.3}} & \cellcolor{gray!6}{\textbf{34.3}} & \cellcolor{gray!6}{\textbf{37.5}} & \cellcolor{gray!6}{\textbf{38.3}} & \cellcolor{gray!6}{\textbf{30.7}} & \cellcolor{gray!6}{\textbf{25.0}} & \cellcolor{gray!6}{16.6} & \cellcolor{gray!6}{10.8} & \cellcolor{gray!6}{1.7} & \cellcolor{gray!6}{3.8}\\
\bottomrule
\end{tabular}
}

\label{tab:competitive-ttechs-imagenet-eps2}
\end{center}
\end{table*}

\begin{table*}
\begin{center}
\caption{Success rate on ImageNet of competitive techniques to train a single surrogate model. Adversarial examples evaluated on ten targets with a maximum perturbation $L_\infty$ norm $\varepsilon$ of $4/255$. Bold is best. In \%.}

\resizebox{\textwidth}{!}{%
\begin{tabular}[t]{lrrrrrrrrrr}
\toprule
\multicolumn{1}{c}{ } & \multicolumn{10}{c}{Target} \\
\cmidrule(l{3pt}r{3pt}){2-11}
Surrogate & RN50 & RN152 & RNX50 & WRN50 & VGG19 & DN201 & IncV1 & IncV3 & ViT B & SwinS\\
\midrule
\cellcolor{gray!6}{Fully Trained SGD} & \cellcolor{gray!6}{44.5} & \cellcolor{gray!6}{25.2} & \cellcolor{gray!6}{24.8} & \cellcolor{gray!6}{27.1} & \cellcolor{gray!6}{16.2} & \cellcolor{gray!6}{16.4} & \cellcolor{gray!6}{9.8} & \cellcolor{gray!6}{8.0} & \cellcolor{gray!6}{1.8} & \cellcolor{gray!6}{3.3}\\
Early Stopped SGD & 51.5 & 27.4 & 27.7 & 28.0 & 18.4 & 18.7 & 10.8 & 10.4 & 2.2 & 2.7\\
\cellcolor{gray!6}{LGV-SWA} & \cellcolor{gray!6}{82.5} & \cellcolor{gray!6}{56.8} & \cellcolor{gray!6}{58.5} & \cellcolor{gray!6}{54.0} & \cellcolor{gray!6}{40.9} & \cellcolor{gray!6}{42.4} & \cellcolor{gray!6}{28.3} & \cellcolor{gray!6}{15.1} & \cellcolor{gray!6}{3.1} & \cellcolor{gray!6}{5.7}\\
SAT  & 76.3 & 62.5 & 66.8 & 63.4 & 48.1 & \textbf{59.0} & \textbf{47.9} & \textbf{40.8} & \textbf{17.4} & \textbf{16.8}\\
\cellcolor{gray!6}{l-SAM \textbf{(ours)}} & \cellcolor{gray!6}{\textbf{85.7}} & \cellcolor{gray!6}{\textbf{70.3}} & \cellcolor{gray!6}{\textbf{73.3}} & \cellcolor{gray!6}{\textbf{73.2}} & \cellcolor{gray!6}{\textbf{58.2}} & \cellcolor{gray!6}{55.6} & \cellcolor{gray!6}{37.9} & \cellcolor{gray!6}{20.5} & \cellcolor{gray!6}{4.0} & \cellcolor{gray!6}{8.2}\\
\bottomrule
\end{tabular}
}

\label{tab:competitive-ttechs-imagenet}
\end{center}
\end{table*}

\begin{table*}[h]
\begin{center}
\caption{Success rate on ImageNet of competitive techniques to train a single surrogate model. Adversarial examples evaluated on ten targets with a maximum perturbation $L_\infty$ norm $\varepsilon$ of $8/255$. Bold is best. In \%.}

\resizebox{\textwidth}{!}{%
\begin{tabular}[t]{lrrrrrrrrrr}
\toprule
\multicolumn{1}{c}{ } & \multicolumn{10}{c}{Target} \\
\cmidrule(l{3pt}r{3pt}){2-11}
Surrogate & RN50 & RN152 & RNX50 & WRN50 & VGG19 & DN201 & IncV1 & IncV3 & ViT B & SwinS\\
\midrule
\cellcolor{gray!6}{Fully Trained SGD} & \cellcolor{gray!6}{77.5} & \cellcolor{gray!6}{52.9} & \cellcolor{gray!6}{51.1} & \cellcolor{gray!6}{55.0} & \cellcolor{gray!6}{33.4} & \cellcolor{gray!6}{36.9} & \cellcolor{gray!6}{21.1} & \cellcolor{gray!6}{15.2} & \cellcolor{gray!6}{3.7} & \cellcolor{gray!6}{6.7}\\
Early Stopped SGD & 82.0 & 56.8 & 54.6 & 59.2 & 35.9 & 41.1 & 24.8 & 18.3 & 3.6 & 5.9\\
\cellcolor{gray!6}{LGV-SWA} & \cellcolor{gray!6}{96.9} & \cellcolor{gray!6}{87.7} & \cellcolor{gray!6}{87.1} & \cellcolor{gray!6}{84.9} & \cellcolor{gray!6}{65.4} & \cellcolor{gray!6}{72.8} & \cellcolor{gray!6}{56.8} & \cellcolor{gray!6}{31.2} & \cellcolor{gray!6}{7.0} & \cellcolor{gray!6}{12.3}\\
SAT & 95.4 & 92.6 & 93.0 & 92.8 & 79.0 & \textbf{90.1} & \textbf{79.1} & \textbf{66.3} & \textbf{38.5} & \textbf{39.1}\\
\cellcolor{gray!6}{l-SAM (ours)} & \cellcolor{gray!6}{\textbf{97.6}} & \cellcolor{gray!6}{\textbf{92.8}} & \cellcolor{gray!6}{\textbf{93.8}} & \cellcolor{gray!6}{\textbf{95.3}} & \cellcolor{gray!6}{\textbf{83.2}} & \cellcolor{gray!6}{85.5} & \cellcolor{gray!6}{71.2} & \cellcolor{gray!6}{42.3} & \cellcolor{gray!6}{9.1} & \cellcolor{gray!6}{19.0}\\
\bottomrule
\end{tabular}
}

\label{tab:competitive-ttechs-imagenet-eps8}
\end{center}
\end{table*}

\begin{table*}[h]
\begin{center}
\caption{Success rate on ImageNet of three complementary categories of transferability techniques evaluated on ten targets with a maximum perturbation $L_\infty$ norm $\varepsilon$ of $2/255$. Underlined is worse when combined with l-SAM. In \%.}

\resizebox{\textwidth}{!}{%
\begin{tabular}[t]{lrrrrrrrrrr}
\toprule
\multicolumn{1}{c}{ } & \multicolumn{10}{c}{Target} \\
\cmidrule(l{3pt}r{3pt}){2-11}
Attack & RN50 & RN152 & RNX50 & WRN50 & VGG19 & DN201 & IncV1 & IncV3 & ViT B & SwinS\\
\midrule
\addlinespace[0.3em]
\multicolumn{11}{l}{\textbf{Model Augmentation Techniques}}\\
\hspace{1em}\cellcolor{gray!6}{GN} & \cellcolor{gray!6}{34.6} & \cellcolor{gray!6}{17.9} & \cellcolor{gray!6}{17.4} & \cellcolor{gray!6}{18.0} & \cellcolor{gray!6}{12.7} & \cellcolor{gray!6}{10.4} & \cellcolor{gray!6}{8.1} & \cellcolor{gray!6}{6.3} & \cellcolor{gray!6}{1.3} & \cellcolor{gray!6}{2.0}\\
\hspace{1em}GN + l-SAM & 59.7 & 42.2 & 42.8 & 45.4 & 35.4 & 29.1 & 18.3 & 11.0 & 1.8 & 2.4\\
\hspace{1em}\cellcolor{gray!6}{SGM} & \cellcolor{gray!6}{26.9} & \cellcolor{gray!6}{14.9} & \cellcolor{gray!6}{15.2} & \cellcolor{gray!6}{15.8} & \cellcolor{gray!6}{15.5} & \cellcolor{gray!6}{9.7} & \cellcolor{gray!6}{7.4} & \cellcolor{gray!6}{6.6} & \cellcolor{gray!6}{1.6} & \cellcolor{gray!6}{3.6}\\
\hspace{1em}SGM + l-SAM & 46.3 & 32.0 & 33.8 & 35.5 & 33.4 & 21.8 & 20.3 & 11.9 & 2.7 & 5.6\\
\hspace{1em}\cellcolor{gray!6}{LGV} & \cellcolor{gray!6}{59.8} & \cellcolor{gray!6}{33.0} & \cellcolor{gray!6}{32.9} & \cellcolor{gray!6}{28.4} & \cellcolor{gray!6}{31.1} & \cellcolor{gray!6}{24.2} & \cellcolor{gray!6}{21.3} & \cellcolor{gray!6}{12.5} & \cellcolor{gray!6}{2.4} & \cellcolor{gray!6}{2.6}\\
\hspace{1em}LGV + l-SAM & \underline{50.7} & \underline{31.3} & 32.9 & 31.4 & 33.5 & 27.9 & 25.0 & 14.3 & \underline{2.1} & \underline{2.5}\\
\addlinespace[0.3em]
\multicolumn{11}{l}{\textbf{Data Augmentation Techniques}}\\
\hspace{1em}\cellcolor{gray!6}{DI} & \cellcolor{gray!6}{46.1} & \cellcolor{gray!6}{27.2} & \cellcolor{gray!6}{30.9} & \cellcolor{gray!6}{30.3} & \cellcolor{gray!6}{22.4} & \cellcolor{gray!6}{24.8} & \cellcolor{gray!6}{17.8} & \cellcolor{gray!6}{15.0} & \cellcolor{gray!6}{2.5} & \cellcolor{gray!6}{4.1}\\
\hspace{1em}DI + l-SAM & 66.6 & 49.5 & 57.1 & 52.3 & 54.1 & 49.3 & 47.5 & 31.8 & 4.4 & 6.9\\
\hspace{1em}\cellcolor{gray!6}{SI} & \cellcolor{gray!6}{26.2} & \cellcolor{gray!6}{14.2} & \cellcolor{gray!6}{14.3} & \cellcolor{gray!6}{13.3} & \cellcolor{gray!6}{10.4} & \cellcolor{gray!6}{11.3} & \cellcolor{gray!6}{8.4} & \cellcolor{gray!6}{7.3} & \cellcolor{gray!6}{0.9} & \cellcolor{gray!6}{1.4}\\
\hspace{1em}SI + l-SAM & 56.5 & 37.9 & 42.9 & 41.2 & 33.0 & 31.4 & 25.0 & 14.7 & 2.1 & 2.9\\
\hspace{1em}\cellcolor{gray!6}{VT} & \cellcolor{gray!6}{26.5} & \cellcolor{gray!6}{14.4} & \cellcolor{gray!6}{14.1} & \cellcolor{gray!6}{13.6} & \cellcolor{gray!6}{10.8} & \cellcolor{gray!6}{10.1} & \cellcolor{gray!6}{6.1} & \cellcolor{gray!6}{6.3} & \cellcolor{gray!6}{1.3} & \cellcolor{gray!6}{2.2}\\
\hspace{1em}VT + l-SAM & 61.5 & 43.0 & 47.0 & 47.4 & 39.4 & 35.9 & 24.3 & 13.3 & 2.1 & 4.9\\
\addlinespace[0.3em]
\multicolumn{11}{l}{\textbf{Attack Optimizers}}\\
\hspace{1em}\cellcolor{gray!6}{MI} & \cellcolor{gray!6}{29.8} & \cellcolor{gray!6}{15.9} & \cellcolor{gray!6}{16.4} & \cellcolor{gray!6}{16.2} & \cellcolor{gray!6}{12.6} & \cellcolor{gray!6}{11.5} & \cellcolor{gray!6}{7.7} & \cellcolor{gray!6}{8.0} & \cellcolor{gray!6}{1.9} & \cellcolor{gray!6}{2.7}\\
\hspace{1em}MI + l-SAM & 58.2 & 41.5 & 45.4 & 44.6 & 39.8 & 35.8 & 28.9 & 17.4 & 2.8 & 5.4\\
\hspace{1em}\cellcolor{gray!6}{NI} & \cellcolor{gray!6}{21.1} & \cellcolor{gray!6}{11.0} & \cellcolor{gray!6}{10.9} & \cellcolor{gray!6}{11.2} & \cellcolor{gray!6}{8.4} & \cellcolor{gray!6}{6.9} & \cellcolor{gray!6}{5.0} & \cellcolor{gray!6}{5.2} & \cellcolor{gray!6}{1.3} & \cellcolor{gray!6}{1.7}\\
\hspace{1em}NI + l-SAM & 44.1 & 28.5 & 30.7 & 32.0 & 25.9 & 19.6 & 11.9 & 9.1 & 1.3 & 2.4\\
\rowcolor{gray!6}
\hspace{1em}RAP & 27.6 & 15.1 & 15.5 & 14.4 & 12.1 & 10.4 & 7.5 & 7.8 & 1.7 & 2.6\\
\hspace{1em}RAP + l-SAM & 54.5 & 38.2 & 42.3 & 41.2 & 37.8 & 33.3 & 27.1 & 16.2 & 3.5 & 6.7 \\
\bottomrule
\end{tabular}
}

\label{tab:table_main_results_complementary_ttechimagenet_eps2}
\end{center}
\end{table*}

\begin{table*}
\begin{center}
\caption{Success rate on ImageNet of three complementary categories of transferability techniques evaluated on ten targets with a maximum perturbation $L_\infty$ norm $\varepsilon$ of $4/255$. Underlined is worse when combined with l-SAM.  In \%.}

\resizebox{\textwidth}{!}{%
\begin{tabular}[t]{lrrrrrrrrrr}
\toprule
\multicolumn{1}{c}{ } & \multicolumn{10}{c}{Target} \\
\cmidrule(l{3pt}r{3pt}){2-11}
Attack & RN50 & RN152 & RNX50 & WRN50 & VGG19 & DN201 & IncV1 & IncV3 & ViT B & SwinS\\
\midrule
\addlinespace[0.3em]
\multicolumn{11}{l}{\textbf{Model Augmentation Techniques}}\\
\hspace{1em}\cellcolor{gray!6}{GN} & \cellcolor{gray!6}{68.0} & \cellcolor{gray!6}{43.1} & \cellcolor{gray!6}{41.3} & \cellcolor{gray!6}{44.1} & \cellcolor{gray!6}{24.8} & \cellcolor{gray!6}{27.2} & \cellcolor{gray!6}{14.3} & \cellcolor{gray!6}{9.9} & \cellcolor{gray!6}{1.9} & \cellcolor{gray!6}{3.8}\\
\hspace{1em}GN + l-SAM & 89.6 & 76.6 & 79.4 & 79.9 & 65.7 & 60.3 & 42.2 & 22.4 & 3.8 & 7.8\\
\hspace{1em}\cellcolor{gray!6}{SGM} & \cellcolor{gray!6}{62.8} & \cellcolor{gray!6}{40.6} & \cellcolor{gray!6}{41.5} & \cellcolor{gray!6}{43.5} & \cellcolor{gray!6}{31.9} & \cellcolor{gray!6}{28.0} & \cellcolor{gray!6}{19.3} & \cellcolor{gray!6}{13.2} & \cellcolor{gray!6}{4.1} & \cellcolor{gray!6}{7.9}\\
\hspace{1em}SGM + l-SAM & 83.2 & 68.7 & 71.5 & 73.0 & 67.0 & 56.2 & 48.9 & 26.6 & 6.2 & 13.6\\
\hspace{1em}\cellcolor{gray!6}{LGV} & \cellcolor{gray!6}{93.3} & \cellcolor{gray!6}{78.1} & \cellcolor{gray!6}{75.3} & \cellcolor{gray!6}{73.1} & \cellcolor{gray!6}{64.4} & \cellcolor{gray!6}{61.6} & \cellcolor{gray!6}{49.3} & \cellcolor{gray!6}{28.8} & \cellcolor{gray!6}{5.0} & \cellcolor{gray!6}{6.5}\\
\hspace{1em}LGV + l-SAM & \underline{88.7} & \underline{74.3} & 75.7 & 75.7 & 70.3 & 61.9 & 56.8 & 31.5 & \underline{4.5} & 7.3\\
\addlinespace[0.3em]
\multicolumn{11}{l}{\textbf{Data Augmentation Techniques}}\\
\hspace{1em}\cellcolor{gray!6}{DI} & \cellcolor{gray!6}{83.1} & \cellcolor{gray!6}{60.5} & \cellcolor{gray!6}{68.1} & \cellcolor{gray!6}{67.3} & \cellcolor{gray!6}{45.4} & \cellcolor{gray!6}{57.9} & \cellcolor{gray!6}{41.4} & \cellcolor{gray!6}{30.7} & \cellcolor{gray!6}{5.7} & \cellcolor{gray!6}{9.9}\\
\hspace{1em}DI + l-SAM & 95.0 & 89.7 & 90.7 & 91.6 & 85.3 & 87.8 & 87.5 & 64.2 & 14.2 & 19.0\\
\hspace{1em}\cellcolor{gray!6}{SI} & \cellcolor{gray!6}{60.0} & \cellcolor{gray!6}{37.9} & \cellcolor{gray!6}{37.3} & \cellcolor{gray!6}{40.0} & \cellcolor{gray!6}{23.9} & \cellcolor{gray!6}{30.0} & \cellcolor{gray!6}{19.6} & \cellcolor{gray!6}{13.5} & \cellcolor{gray!6}{2.6} & \cellcolor{gray!6}{3.8}\\
\hspace{1em}SI + l-SAM & 89.2 & 76.6 & 80.1 & 79.1 & 65.2 & 69.8 & 58.0 & 35.8 & 5.0 & 8.5\\
\hspace{1em}\cellcolor{gray!6}{VT} & \cellcolor{gray!6}{58.6} & \cellcolor{gray!6}{35.0} & \cellcolor{gray!6}{35.2} & \cellcolor{gray!6}{38.5} & \cellcolor{gray!6}{23.9} & \cellcolor{gray!6}{24.7} & \cellcolor{gray!6}{14.9} & \cellcolor{gray!6}{11.0} & \cellcolor{gray!6}{2.3} & \cellcolor{gray!6}{4.9}\\
\hspace{1em}VT + l-SAM & 92.0 & 81.2 & 82.4 & 82.9 & 72.3 & 72.3 & 56.7 & 33.6 & 7.0 & 13.5\\
\addlinespace[0.3em]
\multicolumn{11}{l}{\textbf{Attack Optimizers}}\\
\hspace{1em}\cellcolor{gray!6}{MI} & \cellcolor{gray!6}{56.8} & \cellcolor{gray!6}{37.4} & \cellcolor{gray!6}{37.5} & \cellcolor{gray!6}{38.9} & \cellcolor{gray!6}{27.0} & \cellcolor{gray!6}{29.3} & \cellcolor{gray!6}{18.4} & \cellcolor{gray!6}{14.6} & \cellcolor{gray!6}{3.5} & \cellcolor{gray!6}{4.8}\\
\hspace{1em}MI + l-SAM & 89.4 & 79.3 & 80.4 & 80.8 & 71.5 & 71.1 & 60.1 & 39.3 & 8.5 & 15.2\\
\hspace{1em}\cellcolor{gray!6}{NI} & \cellcolor{gray!6}{53.7} & \cellcolor{gray!6}{33.1} & \cellcolor{gray!6}{32.9} & \cellcolor{gray!6}{35.1} & \cellcolor{gray!6}{20.5} & \cellcolor{gray!6}{20.8} & \cellcolor{gray!6}{12.2} & \cellcolor{gray!6}{9.4} & \cellcolor{gray!6}{1.8} & \cellcolor{gray!6}{3.9}\\
\hspace{1em}NI + l-SAM & 83.9 & 67.3 & 69.8 & 71.4 & 56.1 & 52.5 & 35.6 & 17.6 & 3.8 & 7.0\\
\rowcolor{gray!6}
\hspace{1em}RAP & 58.1 & 36.9 & 36.3 & 39.9 & 26.0 & 27.0 & 15.9 & 13.3 & 3.0 & 5.7\\
\hspace{1em}RAP + l-SAM & 86.0 & 74.5 & 75.4 & 75.5 & 68.7 & 66.5 & 54.7 & 35.6 & 7.9 & 17.1 \\
\bottomrule
\end{tabular}
}

\label{tab:table_main_results_complementary_ttechimagenet}
\end{center}
\end{table*}

\begin{table*}[h]
\begin{center}
\caption{Success rate on ImageNet of three complementary categories of transferability techniques evaluated on ten targets with a maximum perturbation $L_\infty$ norm $\varepsilon$ of $8/255$. Underlined is worse when combined with l-SAM. In \%.}

\resizebox{\textwidth}{!}{%
\begin{tabular}[t]{lrrrrrrrrrr}
\toprule
\multicolumn{1}{c}{ } & \multicolumn{10}{c}{Target} \\
\cmidrule(l{3pt}r{3pt}){2-11}
Attack & RN50 & RN152 & RNX50 & WRN50 & VGG19 & DN201 & IncV1 & IncV3 & ViT B & SwinS\\
\midrule
\addlinespace[0.3em]
\multicolumn{11}{l}{\textbf{Model Augmentation Techniques}}\\
\hspace{1em}\cellcolor{gray!6}{GN} & \cellcolor{gray!6}{92.0} & \cellcolor{gray!6}{73.3} & \cellcolor{gray!6}{69.7} & \cellcolor{gray!6}{74.5} & \cellcolor{gray!6}{45.8} & \cellcolor{gray!6}{50.4} & \cellcolor{gray!6}{29.8} & \cellcolor{gray!6}{19.2} & \cellcolor{gray!6}{3.2} & \cellcolor{gray!6}{7.1}\\
\hspace{1em}GN + l-SAM & 98.2 & 96.5 & 96.5 & 97.4 & 87.3 & 88.3 & 74.4 & 42.9 & 9.0 & 19.4\\
\hspace{1em}\cellcolor{gray!6}{SGM} & \cellcolor{gray!6}{91.2} & \cellcolor{gray!6}{78.4} & \cellcolor{gray!6}{76.2} & \cellcolor{gray!6}{79.2} & \cellcolor{gray!6}{65.1} & \cellcolor{gray!6}{59.7} & \cellcolor{gray!6}{48.2} & \cellcolor{gray!6}{29.1} & \cellcolor{gray!6}{8.9} & \cellcolor{gray!6}{19.6}\\
\hspace{1em}SGM + l-SAM & 97.3 & 95.1 & 96.4 & 96.5 & 91.5 & 88.7 & 84.8 & 59.8 & 18.9 & 32.8\\
\hspace{1em}\cellcolor{gray!6}{LGV} & \cellcolor{gray!6}{99.6} & \cellcolor{gray!6}{97.4} & \cellcolor{gray!6}{95.9} & \cellcolor{gray!6}{95.7} & \cellcolor{gray!6}{87.7} & \cellcolor{gray!6}{91.7} & \cellcolor{gray!6}{79.9} & \cellcolor{gray!6}{47.9} & \cellcolor{gray!6}{8.9} & \cellcolor{gray!6}{16.4}\\
\hspace{1em}LGV + l-SAM & \underline{99.0} & \underline{96.5} & 96.2 & 96.7 & 90.7 & \underline{91.0} & 85.7 & 53.8 & 9.5 & 17.7\\
\addlinespace[0.3em]
\multicolumn{11}{l}{\textbf{Data Augmentation Techniques}}\\
\hspace{1em}\cellcolor{gray!6}{DI} & \cellcolor{gray!6}{96.1} & \cellcolor{gray!6}{90.7} & \cellcolor{gray!6}{91.8} & \cellcolor{gray!6}{91.4} & \cellcolor{gray!6}{74.1} & \cellcolor{gray!6}{88.1} & \cellcolor{gray!6}{72.4} & \cellcolor{gray!6}{55.0} & \cellcolor{gray!6}{14.2} & \cellcolor{gray!6}{20.4}\\
\hspace{1em}DI + l-SAM & 99.8 & 99.6 & 99.5 & 99.7 & 98.6 & 99.3 & 98.4 & 90.4 & 34.7 & 48.7\\
\hspace{1em}\cellcolor{gray!6}{SI} & \cellcolor{gray!6}{90.4} & \cellcolor{gray!6}{70.0} & \cellcolor{gray!6}{69.9} & \cellcolor{gray!6}{71.9} & \cellcolor{gray!6}{47.8} & \cellcolor{gray!6}{60.2} & \cellcolor{gray!6}{42.6} & \cellcolor{gray!6}{29.3} & \cellcolor{gray!6}{6.5} & \cellcolor{gray!6}{10.3}\\
\hspace{1em}SI + l-SAM & 98.9 & 97.3 & 97.3 & 98.0 & 89.9 & 94.8 & 90.1 & 67.2 & 15.0 & 23.6\\
\hspace{1em}\cellcolor{gray!6}{VT} & \cellcolor{gray!6}{79.6} & \cellcolor{gray!6}{62.8} & \cellcolor{gray!6}{61.1} & \cellcolor{gray!6}{63.5} & \cellcolor{gray!6}{41.9} & \cellcolor{gray!6}{48.4} & \cellcolor{gray!6}{32.2} & \cellcolor{gray!6}{23.3} & \cellcolor{gray!6}{6.3} & \cellcolor{gray!6}{10.5}\\
\hspace{1em}VT + l-SAM & 98.0 & 96.7 & 96.1 & 97.3 & 92.9 & 93.1 & 87.1 & 64.2 & 20.0 & 39.2\\
\addlinespace[0.3em]
\multicolumn{11}{l}{\textbf{Attack Optimizers}}\\
\hspace{1em}\cellcolor{gray!6}{MI} & \cellcolor{gray!6}{83.3} & \cellcolor{gray!6}{60.9} & \cellcolor{gray!6}{63.3} & \cellcolor{gray!6}{64.3} & \cellcolor{gray!6}{48.7} & \cellcolor{gray!6}{53.8} & \cellcolor{gray!6}{39.2} & \cellcolor{gray!6}{30.4} & \cellcolor{gray!6}{7.4} & \cellcolor{gray!6}{11.7}\\
\hspace{1em}MI + l-SAM & 98.5 & 96.3 & 96.7 & 97.1 & 91.9 & 92.7 & 88.1 & 68.6 & 21.3 & 31.5\\
\hspace{1em}\cellcolor{gray!6}{NI} & \cellcolor{gray!6}{86.2} & \cellcolor{gray!6}{65.3} & \cellcolor{gray!6}{65.1} & \cellcolor{gray!6}{70.3} & \cellcolor{gray!6}{43.6} & \cellcolor{gray!6}{47.1} & \cellcolor{gray!6}{28.7} & \cellcolor{gray!6}{19.6} & \cellcolor{gray!6}{4.7} & \cellcolor{gray!6}{8.3}\\
\hspace{1em}NI + l-SAM & 97.9 & 94.0 & 95.0 & 96.0 & 87.3 & 86.2 & 74.2 & 42.6 & 10.7 & 21.0\\
\rowcolor{gray!6}
\hspace{1em}RAP & 78.1 & 60.1 & 59.1 & 63.1 & 45.3 & 47.4 & 34.0 & 24.0 & 5.6 & 10.9 \\
\hspace{1em}RAP + l-SAM & 96.3 & 93.3 & 92.5 & 94.2 & 89.5 & 87.4 & 81.0 & 61.0 & 17.9 & 33.6 \\
\bottomrule
\end{tabular}
}

\label{tab:table_main_results_complementary_ttechimagenet_eps8}
\end{center}
\end{table*}

\end{document}